\documentclass[11pt]{article}
\usepackage{emnlp2016}
\usepackage{times}
\usepackage{url}
\usepackage{latexsym}
\usepackage{xspace}
\usepackage{booktabs}
\usepackage{color}
\usepackage{graphicx}
\usepackage{tabulary}
\usepackage{enumitem}
\usepackage{amsfonts}

  {\begin{itemize}[topsep=0pt, partopsep=0pt] %
    \setlength{\itemsep}{0pt}%
    \setlength{\parskip}{0pt}%
    }%
  {\end{itemize}}
  \usepackage[font=small]{caption}  % !
\usepackage[font=small]{subcaption}  % !
  \usepackage{color}
  {\begin{enumerate}≈ \footnotesize%
    \setlength{\itemsep}{0pt}%
    \setlength{\parskip}{0pt}%
    }%
  {\end{enumerate}}

\usepackage[dvipsnames]{xcolor}
\usepackage{subcaption}

\usepackage{microtype}
\usepackage{multirow}
\usepackage{verbatim}
\usepackage{amsmath,amsthm,amssymb}
\usepackage{array}
\usepackage[scaled=0.86]{helvet}
\usepackage{ifthen}
\usepackage{courier}
\usepackage[linesnumbered,vlined,ruled]{algorithm2e}
 \usepackage{lipsum}
\usepackage[export]{adjustbox}

\newcommand{\captionfonts}{\small}
\makeatletter  % Allow the use of @ in command names
\long\def\@makecaption#1#2{%
  \vskip\abovecaptionskip
  \sbox\@tempboxa{{\captionfonts #1: #2}}%
  \ifdim \wd\@tempboxa >\hsize
    {\captionfonts #1: #2\par}
  \else
    \hbox to\hsize{\hfil\box\@tempboxa\hfil}%
  \fi
  \vskip\belowcaptionskip}
\makeatother   % Cancel the effect of \makeatletter

\belowdisplayskip \abovedisplayskip

\usepackage{booktabs}
\usepackage{subcaption}
\usepackage{lipsum}

\emnlpfinalcopy

\newcommand{\todo}[1]{\textcolor{red}{#1}}

\title{Understanding Neural Networks through Representation Erasure}
\author{Jiwei Li, Will Monroe and Dan Jurafsky\\
Computer Science Department, Stanford University, Stanford, CA, USA \\
{\tt jiweil,wmonroe4,jurafsky@stanford.edu} 
}
\date{}

\begin{document}
\maketitle

\begin{abstract}
While neural networks have been successfully
applied to many natural language processing tasks, 
they come at the cost of interpretability. 
In this paper, we propose a general methodology to analyze and interpret 
decisions from a neural model by observing the effects on the model of erasing various
parts of the representation, such as 
input word-vector dimensions, intermediate hidden units, or input words. 
We present several approaches to analyzing the effects of such erasure,
from computing its impact on evaluation metrics,
to using reinforcement learning to erase the minimum set of input words in 
order to flip a neural model's decision. 
In a comprehensive analysis of multiple NLP tasks from lexical
(word shape, morphology) to
sentence-level (sentiment)
to document level (sentiment aspect),
we show that the proposed methodology not only offers clear explanations about neural model decisions, but also provides a way to conduct error analysis on neural models. 
\end{abstract}
\section{Introduction}
A long-standing criticism of neural network models is their lack of interpretability. 
Unlike traditional models that optimize weights on human interpretable features, 
neural network models operate like a black box: using vector representations (as opposed to human-interpretable features) to represent 
text inputs, and applying multiple layers of non-linear transformations.
Mystery exists at all levels of a neural model: At  input layers, what does each word vector dimension stand for? 
What do hidden units in intermediate levels stand for? How does the model
combine meaning from different parts of the sentence, filtering the informational wheat from the chaff? 
How is the final decision made at the output layer?
These mysteries make it hard to tell when and why a neural model makes mistakes, namely, to perform error analysis. This difficulty hinders further efforts to correct these mistakes. 

In this paper, we propose a general methodology for interpreting
neural network behavior by analyzing the effect of erasing pieces of the representation,
to see how such changes affect a neural model's decisions. 
By analyzing the harm this erasure does, we can identify important representations 
that significantly contribute
 to a model's decision; 
by analyzing the benefit this erasure introduces, namely, the cases in which the removal of a representation actually improves a model's decision, we can  identify representations that a neural model inappropriately focuses its attention on, as a form of error analysis.

This erasure can be performed on various levels of representation, 
including input word-vector dimensions, input words or phrases, and
intermediate hidden units.
We apply algorithms of varying complexity for performing this erasure and
analyzing the output. Most simply, we can
directly compute the 
difference in log likelihood on gold-standard labels when representations are erased;
on the more sophisticated end, we offer a
reinforcement learning model to find the minimal set of words that must be erased
to change the model's decision.

The proposed framework offers interpretable explanations for various aspects of neural models:
(1) how a neural model picks word-vector dimensions for linguistic feature classification (parts of speech, named entity recognition, chunking, etc.); (2) how neural models select and filter important words, phrases, and sentences  in sentiment analysis; (3) why architectures like long short-term memory networks (LSTMs) perform more competitively than standard recurrent neural networks (RNNs).
Most importantly, it provides an efficient and general tool to conduct error analysis that can be used on different neural architectures across various NLP applications, which has potential to improve the effectiveness of a wide variety of NLP systems.  

\begin{comment} 
With these models, we are
 able to interpret various fundamental aspects of neural network models in various tasks. For example, (1) to discover the correspondence between word-vector dimensions and 
human-interpretable 
linguistic features (e.g., POS, word shape, chunking, etc); (2) to learn how significantly a particular word or a phrase (represented by a parse tree node) affect sentiment analysis decisions;  
and (3) to measure the contribution of individual words to the overall semantic of a sentence by evaluating the harm it does to downstream semantic evaluation tasks (e.g., the perplexity of predicting the next sentence like the skip-thought model \cite{kiros2015skip}).
\end{comment}
\begin{comment}
The rest of this paper is organized as follows: we detail related work in Section 2. 
We separately describe three individual tasks in 
i.e.,  linking word vector dimensions to linguistic features, rationalize sentiment analysis decision and visualizing the word importance in sentences, respectively described in Sections 4, 5, 6. We briefly conclude this paper in Section 7.  
\end{comment}
\section{Related Work}
Efforts to 
understand  neural vector space models in natural language processing
(NLP)  occur in the earliest work, in which embeddings were visualizing by low-dimensional projection
\newcite{elman89}. Recent work includes
 visualizing 
 state activation \cite{hermans2013training,karpathy2015visualizing}, 
 interpreting semantic dimensions by presenting humans with a list of words
and asking them to choose outliers \cite{fyshe2015compositional,murphy2012learning},
linking dimensions with semantic lexicons or word properties \cite{faruqui2014retrofitting,herbelot2015building},
learning sparse interpretable word vectors \cite{faruqui2015sparse},
and empirical studies of LSTM components \cite{greff2015lstm,chung2014empirical}.

Each of these approaches successfully 
reveals a particular aspect of neural network decisions that is necessary for
 understanding, but each is also constrained by the scope of its applicability.
\newcite{karpathy2015visualizing} visualize the neural generation models  from
an error-analysis point of view, by analyzing predictions
and errors from a recurrent neural models. The approach shows the intriguing dynamics of hidden cells in LSTMs but is limited to a few manually-inspected cases such as brace opening and closing. 
 \newcite{li2015visualizing} use the first-order derivative to examine the saliency of input features, but they rely on the overly strong assumption that the decision score is a linear combination of input features. 

Other closely related work includes that of 
\newcite{rajesh09} and
\newcite{aubakirova2016interpreting}, who
showed how to study unit activations (in autoencoders and CNNs, respectively)
to discover novel features/word clusters.
\newcite{lei2016rationalizing} 
train a separate  
generator that extracts
a subset of text which lead to a similar decision to the original input
 to form an interpretable summary. \newcite{shineural} study the role of 
vector dimensions (for example to track sequence length) in sequence generation tasks;.
\newcite{strobelt2016visual} 
develop an interactive system that 
allows users to select LSTM intermediate states and align these state changes to domain specific structural annotations. 
\newcite{kadar2016representation} 
propose  methods for analyzing the activation patterns of RNNs from a linguistic point of view.
Methods for interpreting and visualizing neural models have also been significantly explored in vision \cite{vondrick2013hoggles,vedaldi2014understanding,zeiler2014visualizing,weinzaepfel2011reconstructing,erhan2009visualizing,simonyan2013deep,kloppel2008automatic}, which we do not describe here for lack of space.

Attention \cite{bahdanau2014neural,luong2015effective,sukhbaatar2015end,rush2015neural,xu2015ask} 
provides an important way to explain the workings of neural models, at least 
for tasks with an alignment modeled between inputs and outputs, like machine translation
or summarization. Representation erasure can be applied to 
attention-based models as well, and can also be used for
tasks that aren't well-modeled by attention.

Our work is also closely related to the idea of adversarial example generation \cite{szegedy2013intriguing,nguyen2015deep}; see
Section \ref{sec-adversarial}. 

%\newcite{szegedy2013intriguing} 
%train a model to fool the network by
%generating adversarial images.
%Similarities and differences between the proposed approach and adversarial example generation will be detailed in 
\begin{comment}
Our work is different from the adversarial example generation work in that adversarial
\end{comment}
\section{Linking Word Vector Dimensions to Linguistic Features}
\label{sec-dimension}
While we know that vector representaitons encode  
aspects of features such as part-of-speech tags and syntactic features
\cite{collobert2011natural}, it is unclear how such
features  are encoded and how tagging models extract the information.

To better understand how these features may be represented, we study
how neural models extract information from word vector dimensions 
make specific classification decisions for widely used linguistic features:
part of speech (POS), named entity class (NER), chunking, prefix, suffix, word-shape and word-frequency. We
first train classifier models on benchmarks with gold-standard labels for these features. 
Then we rationalize a model's decision by analyzing 
the effect of erasure  of input word vectors and of intermediate hidden units.

\begin{figure*}[!ht]
\centering
\begin{tabular}{ccc}
  \begin{subfigure}[t]{0.32\textwidth}
    \includegraphics[width=2in]{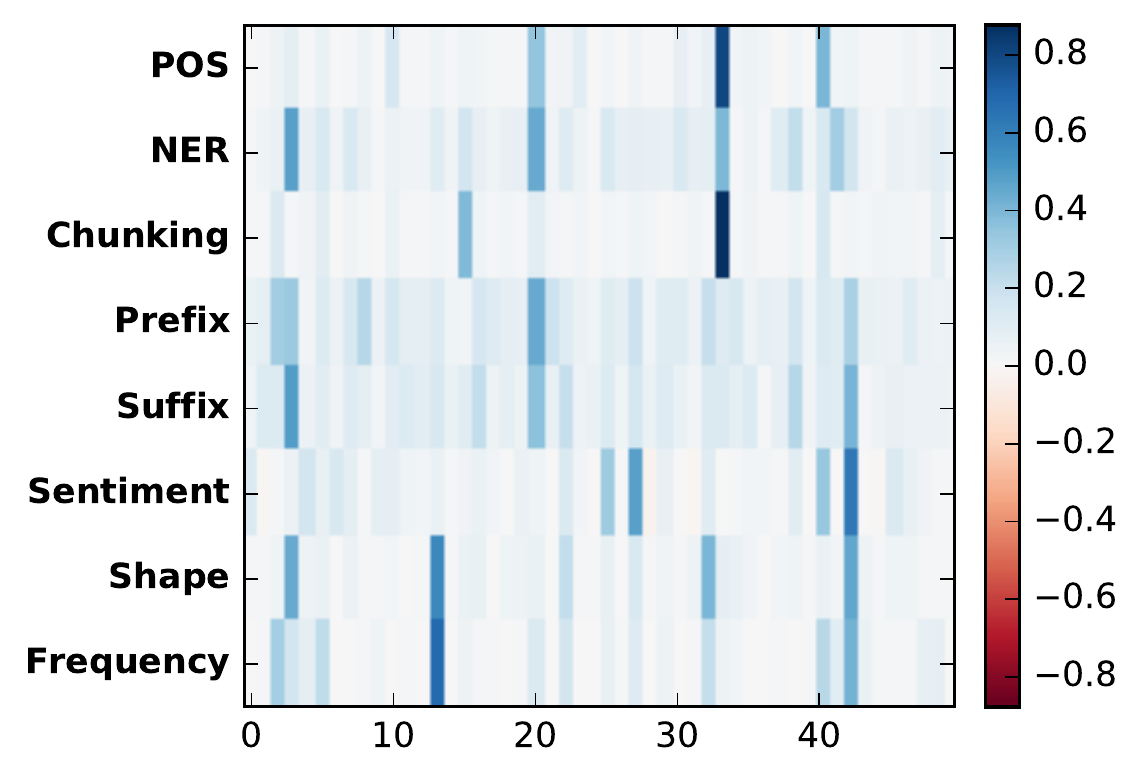}
    \caption{Word2vec, no dropout.}
    \label{dimension-result:w2vnodrop}
  \end{subfigure} &
  \begin{subfigure}[t]{0.32\textwidth}
    \includegraphics[width=2in]{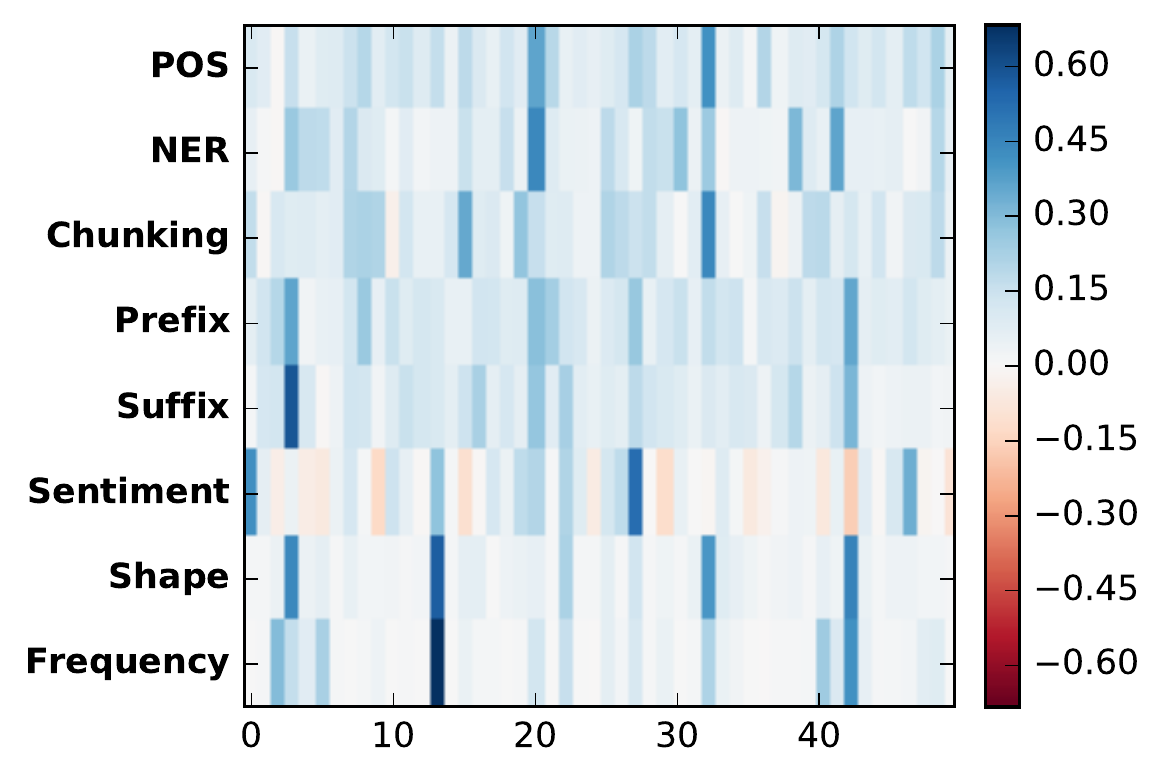}
    \caption{Word2vec, with dropout.}
    \label{dimension-result:w2vdrop}
  \end{subfigure} &
  \begin{subfigure}[t]{0.32\textwidth}
    \includegraphics[width=2in]{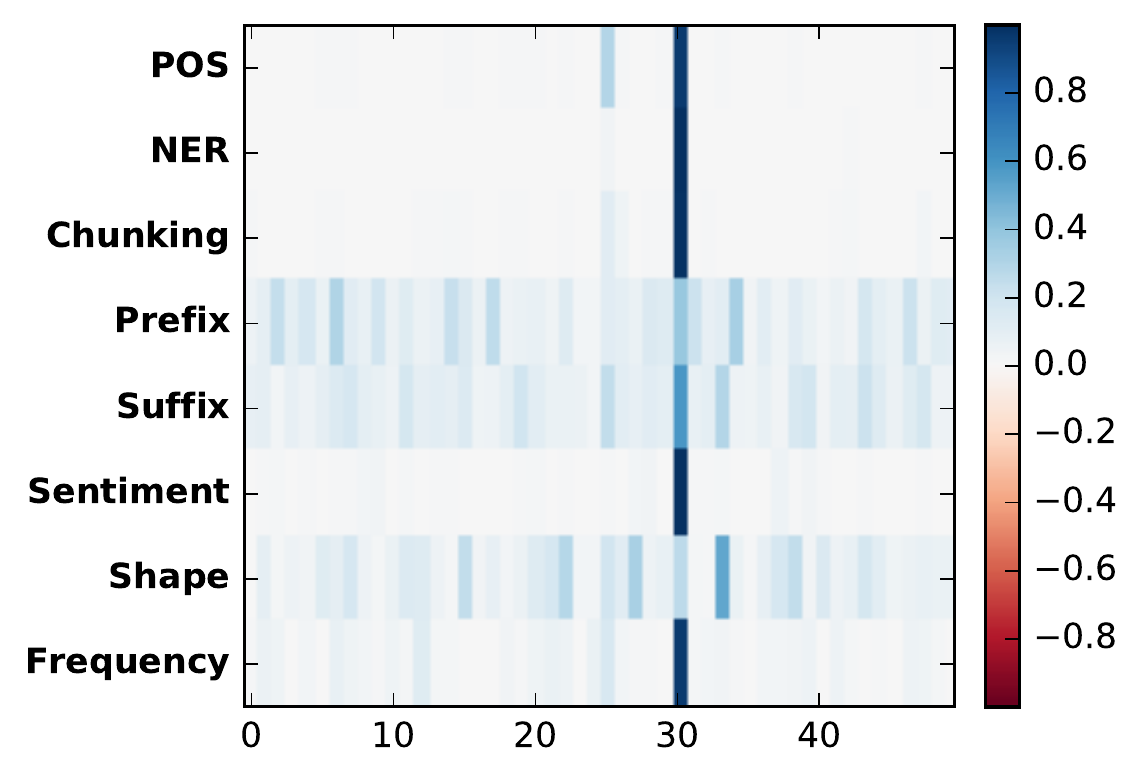}
    \caption{GloVe, no dropout.}
    \label{dimension-result:gnodrop}
  \end{subfigure} \\
  \begin{subfigure}[t]{0.32\textwidth}
    \includegraphics[width=2in]{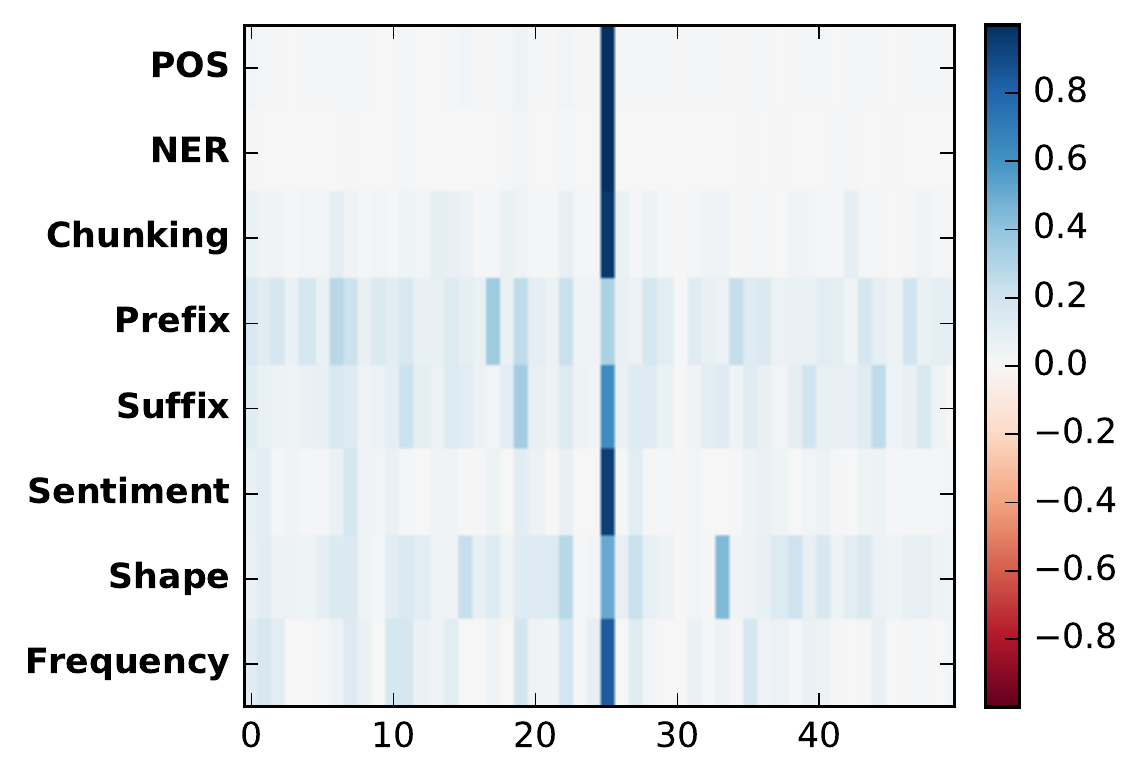}
    \caption{GloVe, no dropout; 31rd dimension removed.}
    \label{dimension-result:gd31}
  \end{subfigure} &
  \begin{subfigure}[t]{0.32\textwidth}
    \includegraphics[width=2in]{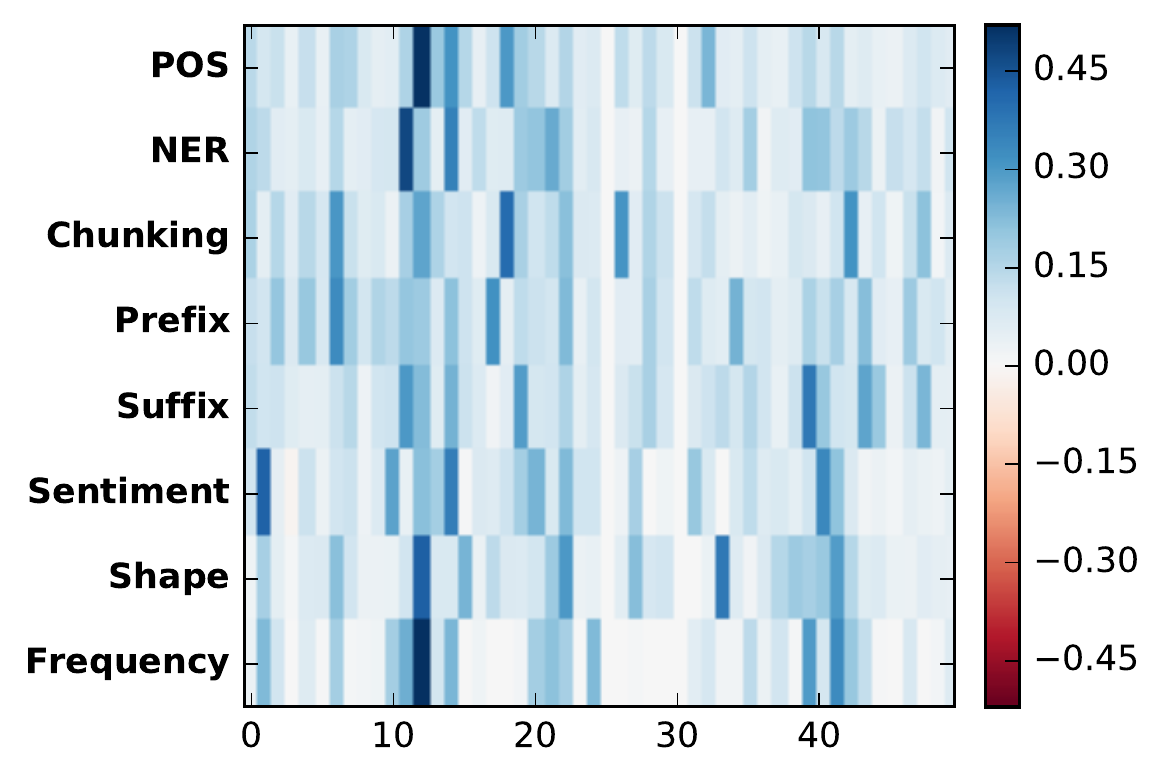}
    \caption{GloVe, no dropout; 31rd, 26th dimensions removed.}
    \label{dimension-result:gd3126}
  \end{subfigure} &
  \begin{subfigure}[t]{0.32\textwidth}
    \includegraphics[width=2in]{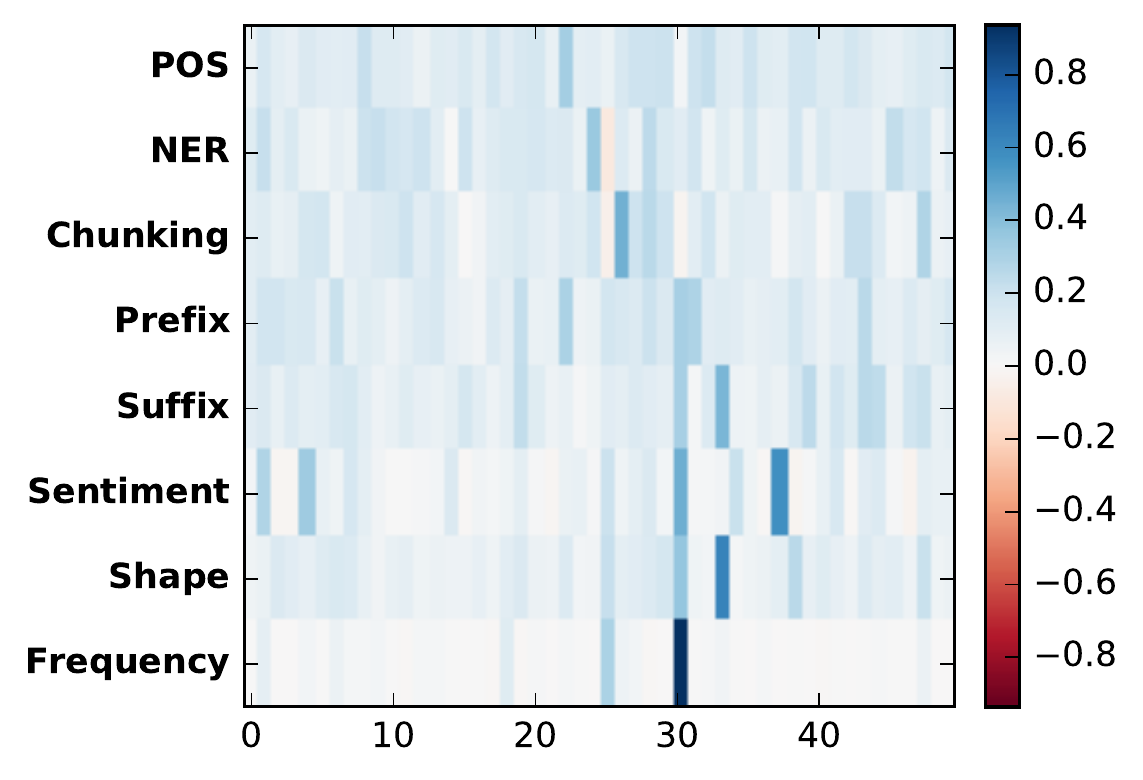}
    \caption{GloVe, with dropout.}
    \label{dimension-result:gdrop}
  \end{subfigure}
\end{tabular}
\caption{Heatmap of word vector dimension importance $I(d)$, computed using Eq.~\ref{di}, for different training strategies and word vectors.  
Each cell shows
the importance of a dimension (column) on each task (row) for the trained model.
Accuracy numbers for each training strategy are shown in Table \ref{test-acc-dimension} in the Appendix.
}
\label{dimension-result}
\end{figure*}

  \subsection{Visualization Model}
  Let $M$ denote a trained neural model.
Given a training example $e\in E$ with gold-standard label $c$, with $L_e$ denoting the index of the tag for $e$, the log-likelihood
assigned by model $M$ to the correct label for $e$ is denoted by
$S(e,c)=-\log P(L_e=c)$. 
Now let $d$ be the index of some vector dimension we are interested in exploring,
and let $S(e,c,\neg d)$ denote the  log-likelihood of 
the correct label for $e$ according to $M$ if
dimension $d$ is erased; that is, its value set to 0.
The \emph{importance} of dimension $d$---denoted by $I(d)$---is the relative difference between $S(e,c)$ and $S(e,c,\neg d)$:
\begin{equation}
I(d)=\frac{1}{|E|}\sum_{e\in E}\frac{S(e,c)-S(e,c,\neg d)}{S(e,c)}
\label{di}
\end{equation}
\subsection{Tasks and Training}
We consider two kinds of tasks: sequence tagging tasks (POS, NER, chunking) and
word ontological classification tasks (prefix, suffix, sentiment, wordshape,
word-frequency prediction); see Appendix Table \ref{Statistics-task} for task details.

%The tasks we consider are divided into two categories:
%(1) Sequence tagging tasks, in which the model needs to assign labels to all tokens within  a sentence. 
%In sequence tagging tasks, 
 %the tag of a word depends on the word itself and its neighbors. 
%The sequence tagging tasks we consider include POS tagging, NER tagging and chunking.
%(2) Word ontology classification tasks, in which no context needs to be considered. The input is just the word itself, and the model needs to assign a label to the input word. 
%Tasks we consider in this category include prefix, suffix, sentiment and word-shape classification as well as (log) word-frequency prediction.
%

For sequence tagging tasks, the input consists of the concatenation of the vector representation of the word to tag and the representations of its neighbors (window size is set to 5). 
For ontology tagging tasks, the input is just the representation of the input word. 
We study  word2vec \cite{mikolov2013distributed} and GloVe \cite{pennington2014glove} vectors, each
50-dimensional vectors pre-trained using the Gigaword-Wiki corpus. 
For each task, we train a four-layer neural model (an input word-embedding layer, 2 intermediate layers, and a output layer that outputs a scalar) using a structure similar to that of \newcite{collobert2011natural} with a \textsc{tanh} activation function.
Each intermediate layer contains 50 hidden units.
Test accuracy for each task is shown in Appendix Table \ref{test-acc-dimension}.

\begin{figure}[t]
\centering
\includegraphics[width=3in]{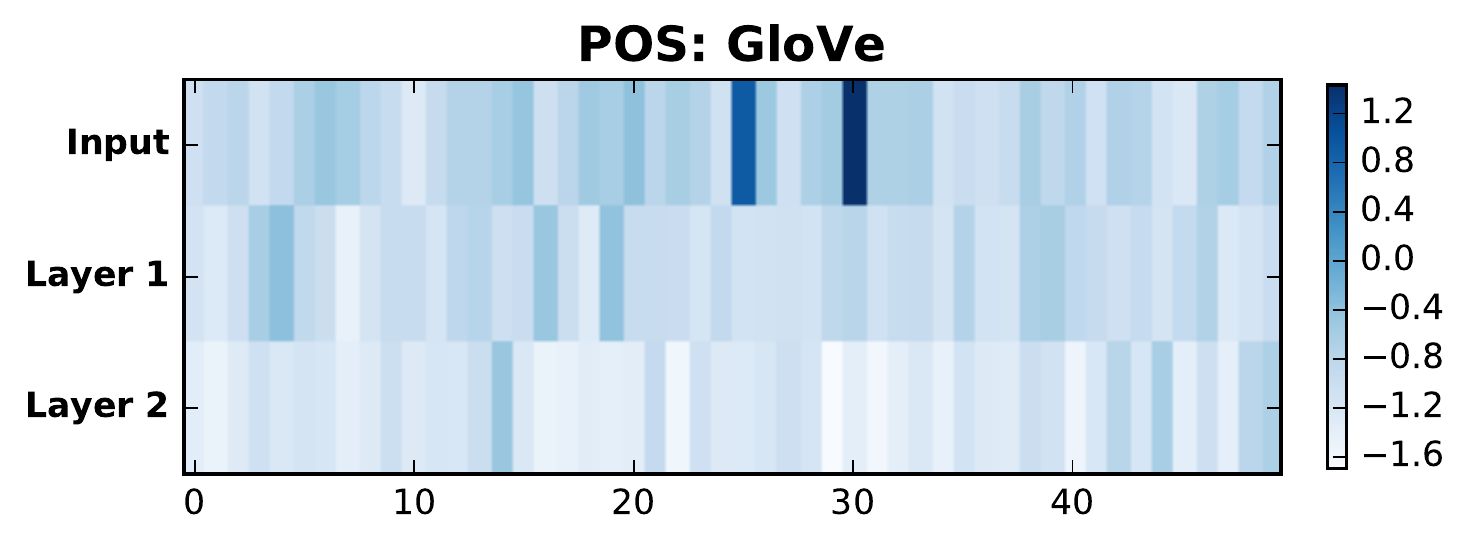}
\includegraphics[width=3in]{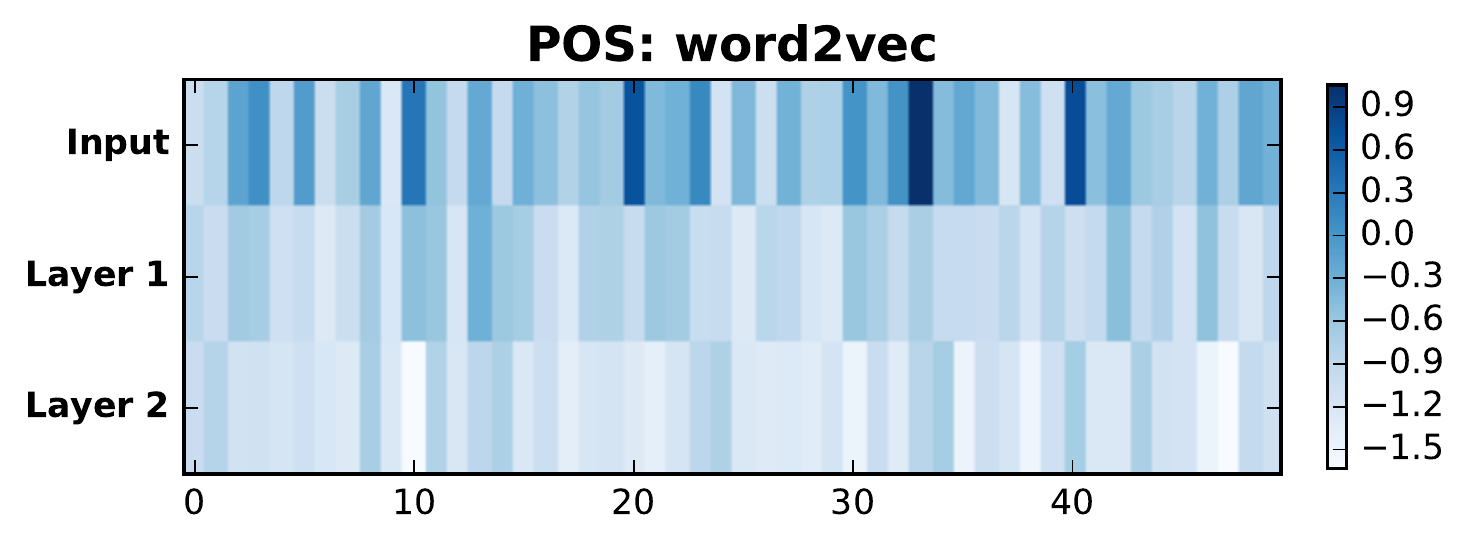}
\caption{Heatmap of importance (computed using Eq.~\ref{di}) of each layer for the POS task. 
Each cell corresponds a unit in a neural model layer. 
Each column denotes a dimension and each row denotes a layer in the network. 
Importance values are projected to log space.}
\label{layer}
\end{figure}

\begin{figure}[t]
\centering
    \includegraphics[width=2.3in]{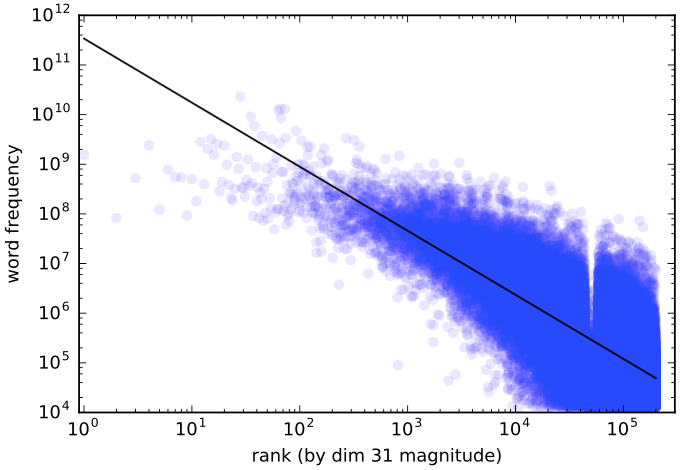}
     \includegraphics[width=2.3in]{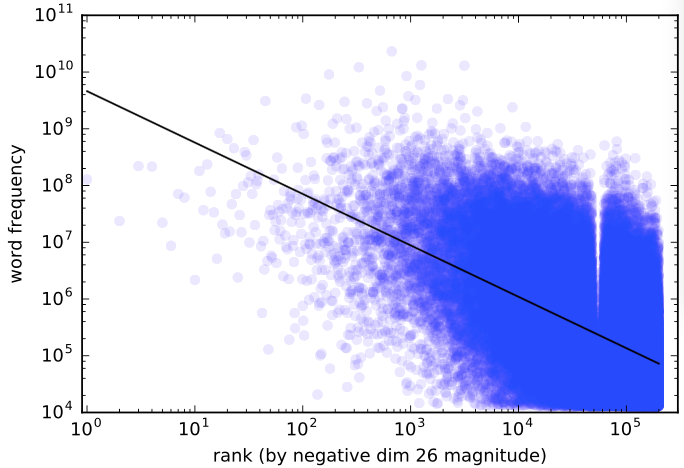}
  \caption{Correlation with word frequency of the magnitude of (a) the 31st dimension ($R^2 = 0.55$, $p < 1\times{}10^{-5}$) and (b) the 26th dimension  ($R^2 = 0.27$, $p < 1\times{}10^{-5}$) of GloVe vectors.}
  \label{rank}
\end{figure}

\subsection{Results}
For each task, we take the pre-trained model, 
erase an input word dimension by setting its value to 0, 
apply the pre-trained model to the modified inputs, and apply Eq.~\ref{di} to compute the importance score of the erased dimension. 

Results are shown in Figure \ref{dimension-result}. 
Each row corresponds to a feature classification task (e.g., POS, NER) and each column in a row signifies 
the importance of a word-vector dimension to the pre-trained model
for that task. 
For word2vec vectors (shown in Figure~\ref{dimension-result}a), we observe clear patterns that
the model focuses more on some dimensions than others and that 
 some tasks share important dimensions. For example, POS and chunking share dimension 34; NER, prefix and suffix share dimensions 4 and 31; etc. 
When applying dropout \cite{srivastava2014dropout}, we can clearly see that importance is distributed more equally among different dimensions, which is intuitive since the model is forced to make use of other dimensions when the dominating dimension is dropped during training. 

Things are a bit more confusing with Glove vectors  (Figure~\ref{dimension-result}c): we observe a single dimension (d31) dominating across almost all tasks. Interestingly, if we remove dimension d31 and retrain the model, another dominant dimension  (d26) appears (Figure~\ref{dimension-result}d). 
Only if we remove both these dimensions 
(Figure~\ref{dimension-result}e) can the model spread its attention to most of the other dimensions. 
Interestingly,
performance does not drop after removing these two dimensions and retraining the models (as shown in Table~\ref{test-acc-dimension} in the Appendix).

In Figure~\ref{dimension-result}f, which shows the effects of using dropout,
the influence of these two dimensions (26 and 31) 
declines dramatically in most  tasks but 
still stands out in frequency regression, suggesting 
that these two dimensions are associated with word frequency.
Indeed,  when we rank words by dimension magnitude, Figure \ref{rank} shows a large
correlation between word frequency and the values of the 26th and 31st dimension.
Our results suggest that 
models trained on GloVe vectors rely on these frequency dimensions
because of the usefulness of word frequency, but manage to get sufficient
information from other redundant dimension when these are eliminated.\footnote{
Word2vec vectors don't contain dimensions strongly associated with frequency, presumably
because tokens are omitted in proportion to word-type frequency in word2vec models \cite{mikolov2013efficient}. 
These differences  may explain the differing suitability of
GloVe and word2vec embeddings for different NLP tasks.}

%Our explanation for the behavior of the model trained on GloVe vectors is as follows: 
%since word frequency is an important feature for many classification tasks, 
 %the trained model tends to attach strong focus to the  frequency-based dimensions. However, dimensions other than the word-frequency dimensions also bear enough information for the model to make correct decisions. This is why when the word-frequency dimensions are eliminated, the model still performs as competitively. 
% It is also worth noting that word2vec vectors don't bear the same strong word-frequency information as GloVe vectors. We conjecture that this is because tokens are omitted in proportion to word-type frequency in word2vec models \cite{mikolov2013efficient}. 
%
\begin{table}
\centering
\scriptsize
\begin{tabular}{cccc}
rank&Bi-LSTM&Uni-LSTM&RNN\\\hline
1& masterpiece (104)&masterpiece (32)&pathetic (8.3)  \\ 
2&sweetest (47)& dreadful (32)&dreadful (6.2)\\
3&dreadful (44)&sweetest (14)&brilliant (5.6)\\
4&stillborn (21)&pathetic (9.8)&ungainly (4.6)\\
5&pathetic (17)&flawless (7.8)&smartest (4.4)\\
6&eye-popping (13)&breathtaking (6.7)&hated (4.3)\\
7&succeeds (13)&dumbness (6.6)&eye-popping (4.1)\\
8&breathtaking (12)&beaut (6.3)&stupider (3.4)\\
9&ugliest (9.8)&disappointingly (6.2)&dicey (3.3)\\
10&flawless  (9.6)&heady (6.1)&masterpiece (3.3)\\\hline
\end{tabular}
\caption{Top 10 ranked words by importance (computed using Eq.~\ref{di}) from the Bi-LSTM, Uni-LSTM and standard RNN models.}
\label{table:top10}
\end{table}

Figure \ref{layer} shows  importance values for hidden unit dimensions in different layers on the POS task (see Appendix Figure~\ref{layer_all_tasks} for other tasks). The heatmap color is generally lighter in the higher layers, meaning that on higher layers
importance is distributed more equally across the dimensions.
% The more equal distribution of importance among higher layer dimensions makes
  %the importance scores  generally smaller in higher layers than lower ones, which means 
%The fact that importance scores are lower in higher layers than lower ones means the
%classification decision is more robust to the change of one particular dimension 
%in higher layers, in keeping with 
%Such a phenomenon fits with our expectation that a 
In other words, neural models tends to distill information from a few important dimensions in the input layer, making the removal of these input layer dimensions more detrimental. At higher layers, however, the information is spread across different units and the importance scores are generally lower,
meaning that the final classification decision is more robust to the change in any particular dimension.
\begin{figure}
\centering
\small
\includegraphics[width=2.7in]{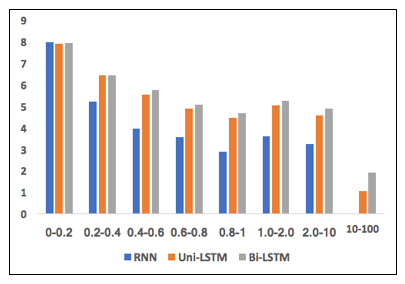}
\caption{Histogram of words by importance for different models. Word frequency is projected to log space.}
\label{Histogram}
\end{figure}

\section{Finding Important Words in Sentiment Analysis}
The section above is concerned mostly  with individual vector dimensions. 
However, for most tasks in NLP, 
 words rather than individual dimensions
function as basic units. In this section, we demonstrate how the proposed model can facilitate the understanding of neural models at the word level. 
In this section  we consider the
Stanford Sentiment Treebank dataset \cite{socher2013recursive}, which focuses on phrase/sentence level classification.

\begin{table*}[ht]
\scriptsize
\begin{tabular}{p{0.7cm}p{2cm}p{1cm}p{1cm}p{10cm}}\\
rank&Word&Score&Label&Original Sentence \\\hline
1&revelatory&-0.90&-&flat, but with a revelatory performance by michelle williams.\\
2&lacks&-0.88&+&what it lacks in originality it makes up for in intelligence and b-grade stylishness.\\
3&shame&-0.84&+&it takes this never-ending confusion and hatred, puts a human face on it, evokes shame among all who are party to it and even promotes understanding.\\
4&skip&-0.83&+&skip work to see it at the first opportunity.\\
5&lackadaisical&-0.82&+&a pleasant ramble through the sort of idoosyncratic terrain that errol morris has often dealt with... it does possess a loose, lackadaisical charm.\\
6&by-the-books&-0.82&+&a fairly by-the-books blend of action and romance with sprinklings of intentional and unintentional comedy.\\
7&misses&-0.82&++&this is cool, slick stuff, ready to quench the thirst of an audience that misses the summer blockbusters.\\
8&bonehead&-0.82&++&the smartest bonehead comedy of the summer.\\
9&dingy&-0.81&+&it's a nicely detailed world of pawns, bishops and kings, of wagers in dingy backrooms or pristine forests.\\
10&enjoying&-0.81&-&i kept thinking over and over again,' i should be enjoying this.'\\
11&foul&-0.80&+&a whole lot foul, freaky and funny.\\
12&best&-0.80&-&the best way to hope for any chance of enjoying this film is by lowering your expectations.\\
25&pleasing&-0.72&-&an intermittently pleasing but mostly routine effort.\\
\hline
\end{tabular}
\caption{Words with high negative importance score (computed using Eq.\ref{di}) obtained by the Bi-LSTM model. Negative importance means that the model makes better prediction when the word is erased. ++, +, 0, -, -- respectively denote strong positive, positive, neutral, negative and strong negative sentiment labels. which are gold-standard ones from the dataset. }
\label{partial-top}
\end{table*}

We can compute the importance of words similarly to that of word-vector dimensions, by calculating the relative change of the log-likelihood of the correct sentiment label for a text unit when a particular word is erased. 
The formula is exactly the same as Eq.~\ref{di}, but with dimensions replaced by words. 

We examine three models:
a {\it standard RNN} with \textsc{tanh} activation functions,
an LSTM ({\it Uni-LSTM}) and a bidirectional LSTM ({\it Bi-LSTM}), all trained on the Stanford treebank dataset. 
We  first transform each parse tree constituent in the dataset to a sequence of tokens. Each sequence is then
 mapped to a phrase/sentence representation 
and fed to a softmax classifier.  
The  {\it Bi-LSTM}, {\it Uni-LSTM} and standard {\it RNN} respectively obtain an  accuracy of 0.526, 0.501 and 0.453 on sentence-level fine-grained classification.
It is worth noting that 
the {\it Bi-LSTM} model achieves state-of-the-art performance
 in sentence-level fine-grained classification on this benchmark, 
 significantly 
 outperforming tree-based 
models, 
namely 50.1 
 reported in \newcite{zhu2015long} and 51.0 in \newcite{tai2015improved}.
We refer the readers to the Appendix 
for more details about the dataset and model training. 
 \begin{figure*}[t]
  \centering
  \begin{tabular}{cc}
    \begin{subfigure}{0.3\textwidth}
      \includegraphics[scale=0.45,right]{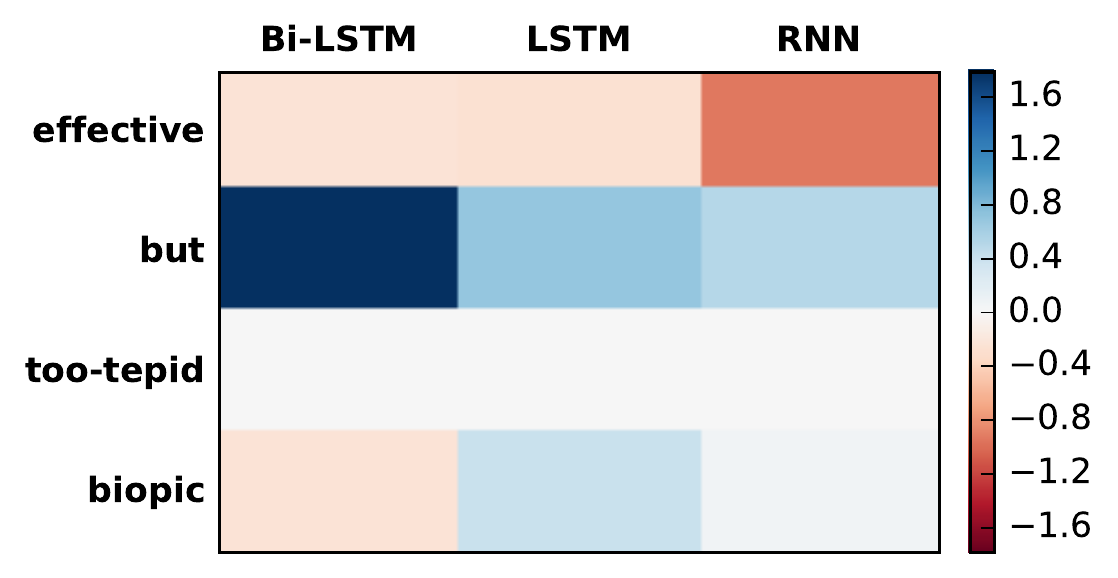}
      \caption{Neutral}
      \label{fig:heat-sentiment:biopic}
    \end{subfigure}
    &
    \begin{subfigure}{0.3\textwidth}
      \includegraphics[scale=0.45,right]{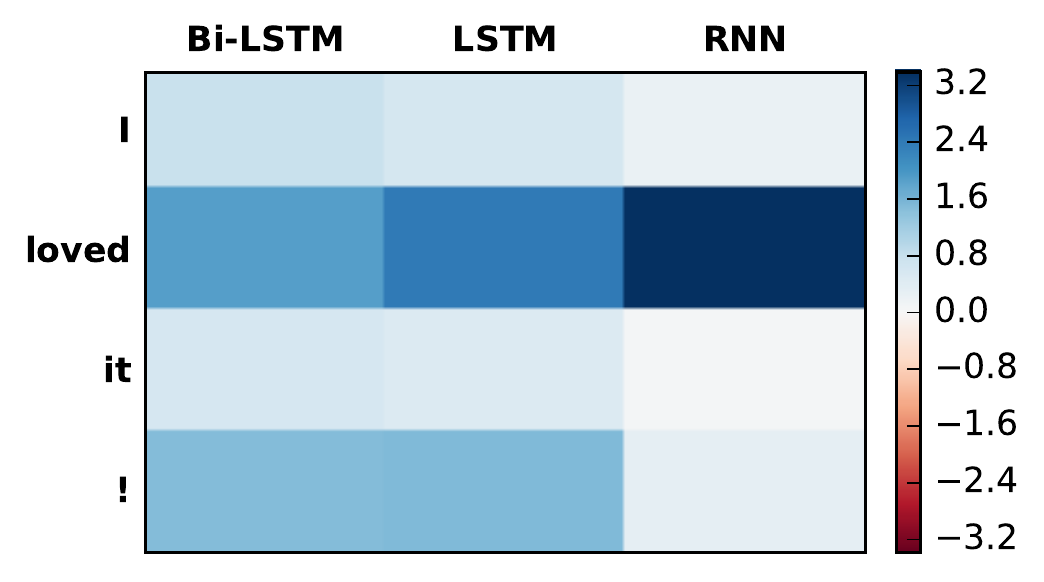}
      \caption{Strong positive}
     \label{fig:heat-sentiment:lovedit}
    \end{subfigure}
    \\
    \begin{subfigure}{0.3\textwidth}
      \rule{0pt}{2ex}
      \includegraphics[scale=0.45,right]{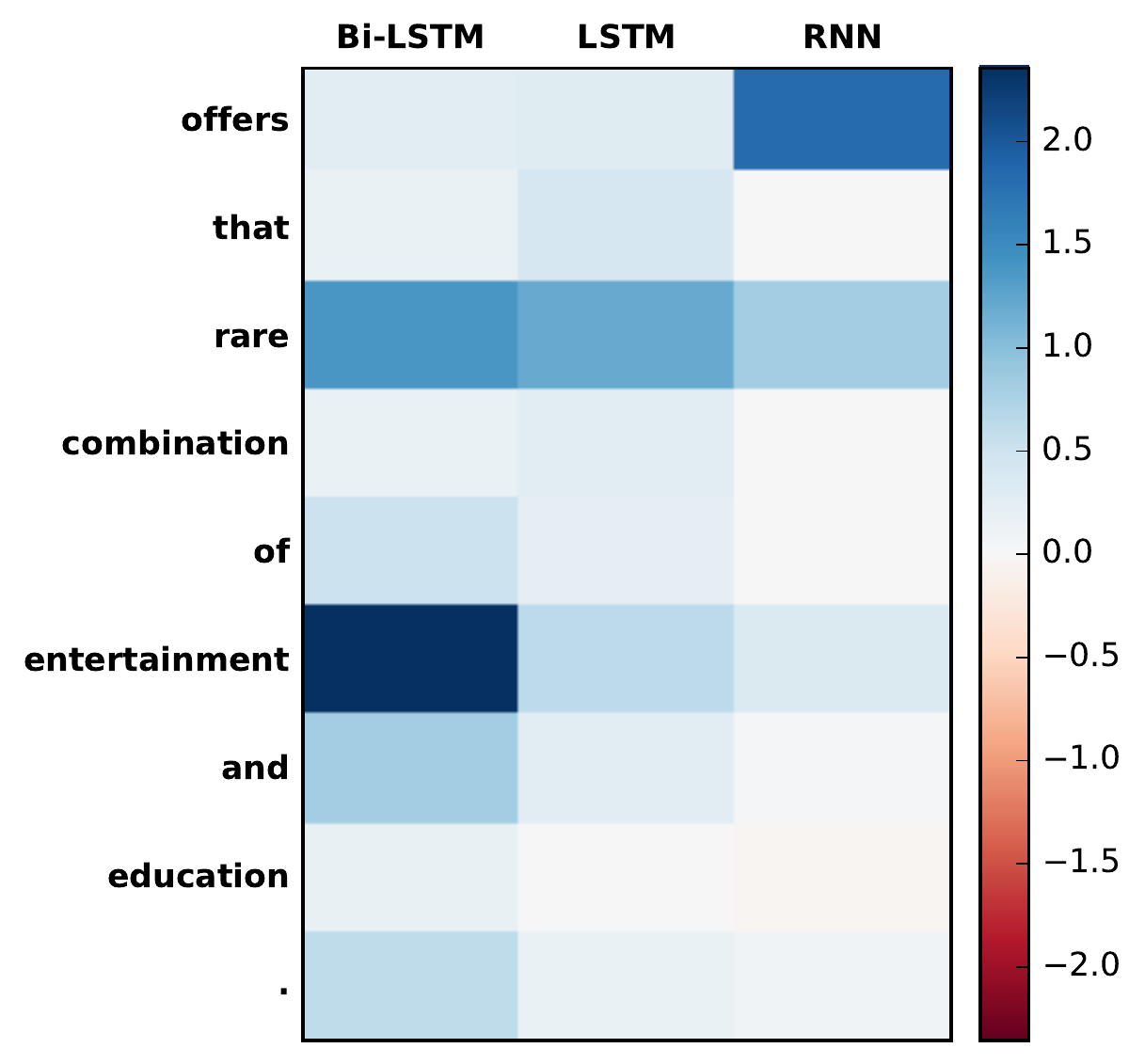}
      \caption{Strong positive}
      \label{fig:heat-sentiment:combination}
    \end{subfigure}
    &
    \begin{subfigure}{0.3\textwidth}
      \rule{0pt}{2ex}
      \includegraphics[scale=0.45,right]{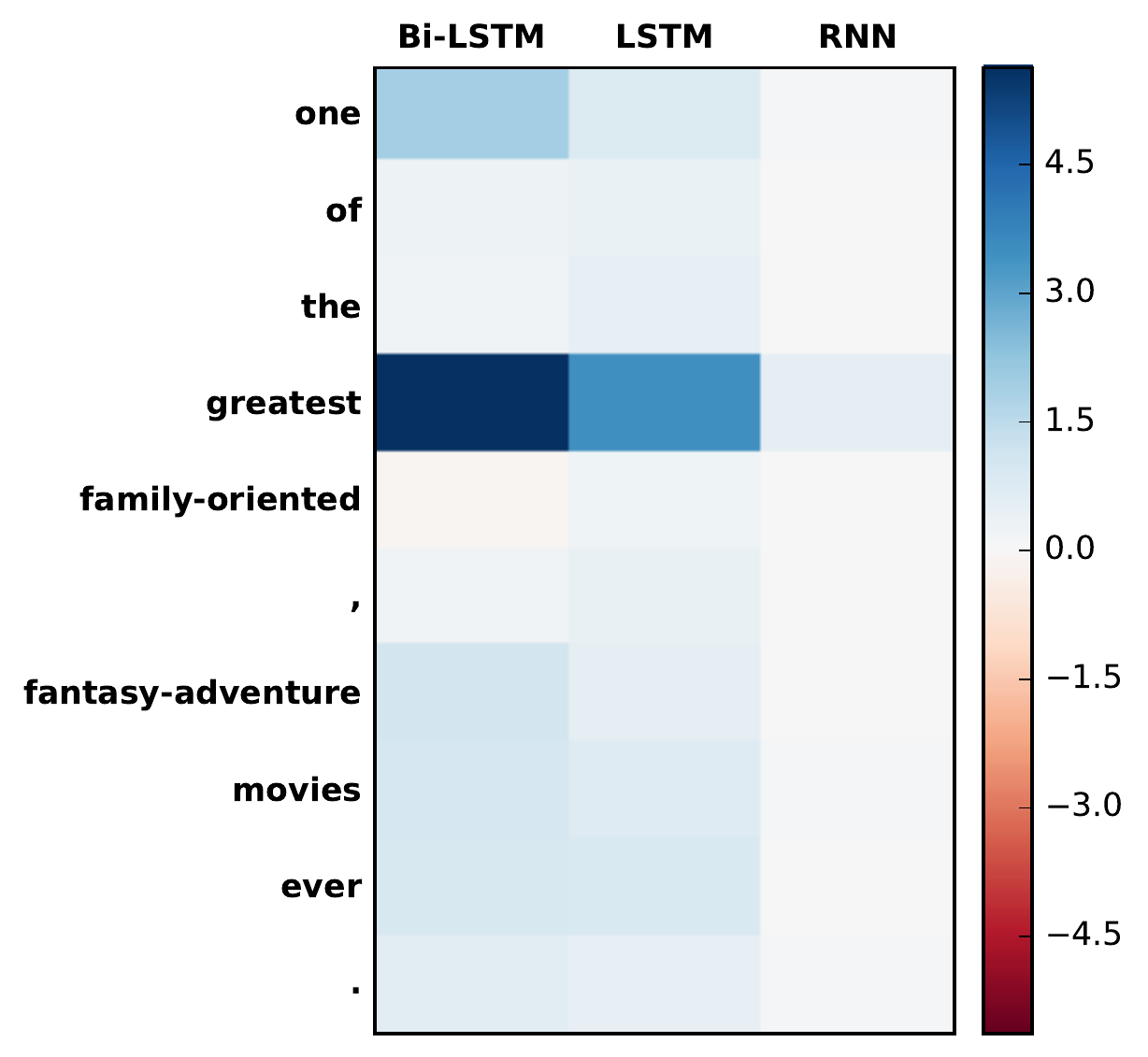}
      \caption{Strong positive}
      \label{fig:heat-sentiment:greatest}
    \end{subfigure}
  \end{tabular}
  \begin{subfigure}{0.3\textwidth}
    \includegraphics[scale=0.525,center]{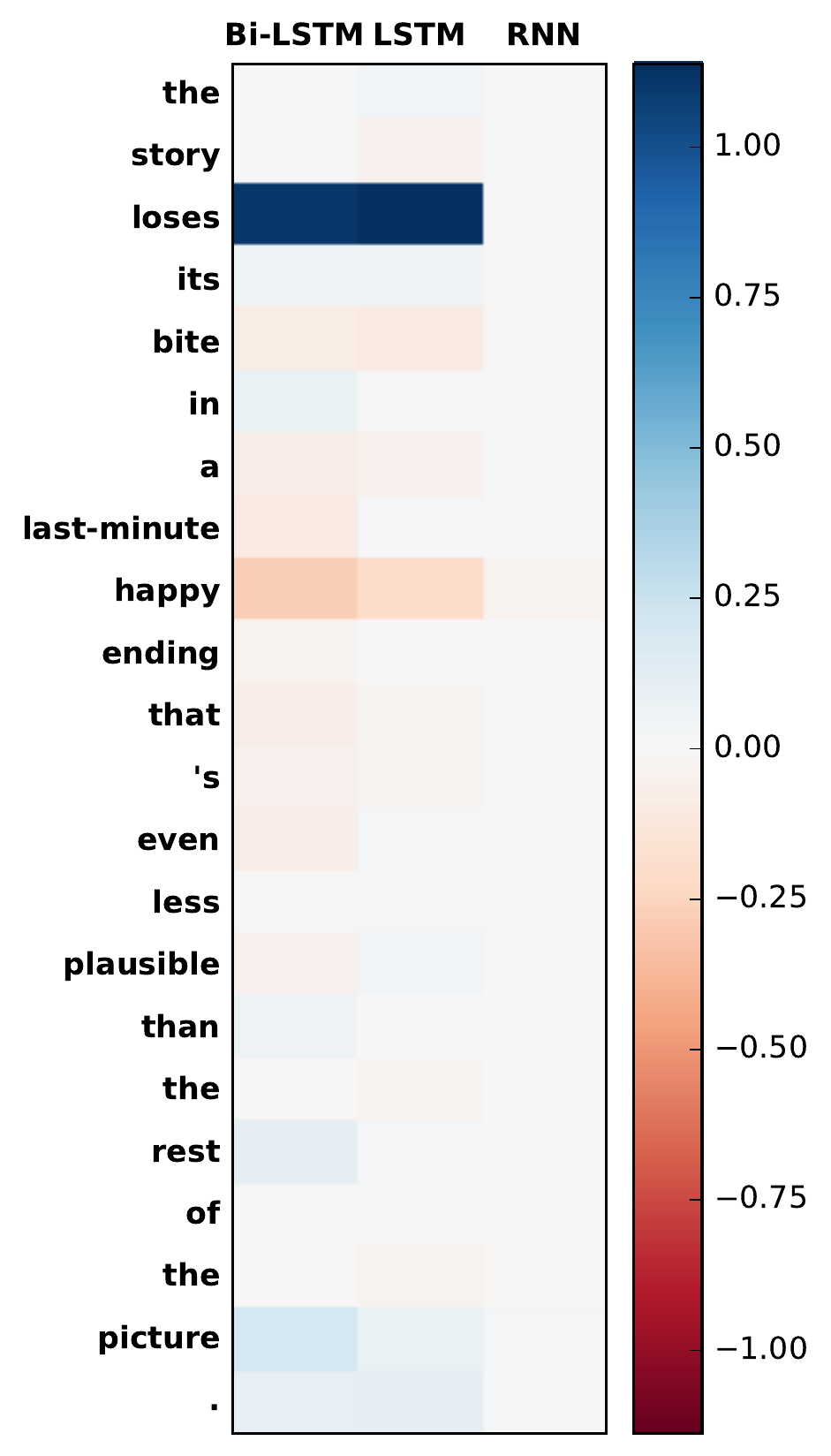}
    \caption{Strong negative}
    \label{fig:heat-sentiment:bite}
  \end{subfigure}
  \caption{Heatmap of word importance (computed using Eq.~\ref{di}) in sentiment analysis.}
   \label{fig:heat-sentiment}
 \end{figure*}

We present
the importance scores of a few selected sentiment-indicative words in Table~\ref{seed}. 
The ranking score is computed by averaging the log-likelihood difference resulting from erasing that word across all test examples containing the word. 
We
can see that 
the {\it Bi-LSTM} is more sensitive to the deletion of these  sentiment indicators than the {\it Uni-LSTM}, which is in turn more sensitive than the {\it RNN}.
This is presumably due to the gate structures
in LSTMs that control information
flow, making these architectures better at
focusing on words that indicate sentiment.

The highest-ranked words by importance  (computed using Eq.\ref{di})
for each model are listed
 in Table~\ref{table:top10} (more comprehensive lists are presented in Table~\ref{all-word-importance} in the Appendix).
 Figure~\ref{Histogram} shows a histogram of all words by importance for  different models. The distribution also confirms that the {\it Bi-LSTM} model is more sensitive to the sentiment-indicative words, with more words in buckets with higher importance values.

\begin{table}
\centering
\small		
\begin{tabular}{cccc}
word&Bi-LSTMs&Uni-LSTMs&RNN\\\hline
greatest&9.463&5.593&0.742 \\
wonderful&9.521&3.292&0.704\\
worst&7.739&4.698&0.967\\
excellent&6.835&4.883&1.859\\
best&4.916& 2.448&0.548\\
hated&6.557&3.512&4.338\\
love&1.678&1.786&0.999 \\
unforgettable&2.286&1.648&1.482\\
waste&4.579&3.600&2.342 \\
disaster&3.728&3.362&0.021\\\hline
\end{tabular}
\caption{Importance score (computed using Eq.\ref{di}) for a few  sentiment indicators assigned by different models. }
\label{seed}
\end{table}

Figure~\ref{fig:heat-sentiment} plots the importance score of individual words (rows) for the different models (columns) in a few specific examples of sentence-level sentiment classification.
Higher values mean that the model is more sensitive to the erasing of a particular word. 
As can be seen, all three models attach more importance to words that are indicative of sentiment (e.g., ``loved'', ``entertainment'', ``greatest'') and  
 dampen the influence of
other tokens. 
LSTM-based models generally show a clearer focus on sentiment words than standard RNN models, and they also succeed in attaching importance to intensification tokens (e.g., the exclamation mark in Figure~\ref{fig:heat-sentiment}b), which the {\it RNN} fails to identify.

We also notice an interesting phenomenon in Figure~\ref{fig:heat-sentiment}: the importance scores of  words can take negative values, which means that the removal of some words actually improves the model's decision. 
Such discoveries can help with  error analysis on a model by identifying which words confuse the model and lead to mistakes. 
We therefore also list the top-ranked words by negative importance score (the removal of which words can best help the model make the correct decision). We present some of the top negative important words obtained using the Bi-LSTM model in Table~\ref{partial-top}, while listing comprehensive results from all three models in Tables~\ref{bi-re}, \ref{uni-re} and \ref{rnn-re} in the Appendix. 
From these tables, we can clearly identify a few patterns that make neural models fail:
(1) A common sentiment indicator word is
used in a context (e.g., describing details of the movie) that makes the word not bear any sentiment 
orientation, such as the word {\it happy} in {\it happy ending} (Figure~\ref{fig:heat-sentiment}e), or
{\it shame} (Table~\ref{partial-top}, rank 3). (2) A sentiment indicator word is used in a specific context 
that turns its sentiment into the opposite of its common usage;
 e.g., ``the smartest bonehead'' (Table~\ref{partial-top}, rank 8).
(3) A sentiment indicator is used in the scope of an irrealis  modal ---e.g., {\it i should be enjoying this} (Table \ref{partial-top}.rank10)---or in an ironic context---e.g., 
{\it the best way to hope for any chance of enjoying this film is by lowering your expectations} (Table \ref{partial-top}, rank 12). 
(4) A sentiment indicator is used in a concessive sentence, requiring the handling of discourse information; e.g., {\it revelatory} in {\it 
flat, but with a revelatory performance by michelle williams} (Table~\ref{partial-top}, rank 1),
{\it pleasing} in 
 {\it an intermittently pleasing but mostly routine effort} (Table~\ref{partial-top}, rank 25).
 Resolving these problems  is a long-term goal of future work in
sentiment analysis.

\section{Reinforcement Learning for Finding Decision-Changing Phrases} \label{sec-adversarial}

The analysis that we have described so far deals with individual words or dimensions.
How can representation erasure help us understand the importance of larger compositional
text units like phrases or sentences? We propose another technique:
removing the minimum number of words to change the model's prediction\footnote{Our technique is 
closely related to adversarial example generation \cite{szegedy2013intriguing,nguyen2015deep}, the idea
of finding the minimal change to input dimensions to change neural network decisions.
It differs in two ways: (1)  adversarial training is usually not suited for interpreting how a model makes a decision, but rather for detecting the intrinsic flaws of the model; these
adversarial examples are usually very similar to real examples (often indistinguishable by humans) but can fool the model into making a different decision. (2) 
Words are a basic unit in NLP; because changing
dimensions may harm text integrity (e.g., break the language model)
our model removes words rather than dimensions,
making our proposed method discrete rather than the continuous method of adversarial example generation.
}.
More formally,
let $e$ denote
an input text unit
consisting of a sequence of words, $e=\{w_1,w_2,...,w_N\}$, where $N$ denotes the number of words in $e$, and  let
$L_e$ denote the index of the label that $M$ gives to $e$. 
 The task is to discover a minimal subset of $e$, denoted by $D\subset e$, such that the removal of all words in $D$ from $e$ (the remaining words are denoted by $e-D$) will change the label $L_e$. Let $|D|$ denote the number of words in D. The problem is formalized as follows
\begin{equation}
\min_D |D|~~~~ s.t. ~~L_{e-D}\neq L_e
\end{equation} 
Finding the optimal solution requires enumerating all different word combinations, which is computationally intractable when the number of words in $e$ gets large. 
To address this issue, 
we propose an  strategy based on reinforcement learning to find an approximate solution.

Given a pre-trained sentiment classification model $M$, an input example $e$, and the label $L_e$ 
that $M$ gives to $e$, 
we define a policy $\pi$ over
a binary variable $z_t$, indicating 
 whether a word $w_t\in e$ should be removed.  $z_t$ takes the value of 1 when $w_t$ is removed and $0$ otherwise. The policy model takes as input the representation associated with word $w$
at the current time step 
outputted from model $M$
and defines a binary distribution $\pi$ over $z_t$. 
The policy model  examines every word in $e$ and decides whether the word should be kept or removed.
Let $D$ be the union of the removed words.
After the policy model finishes removing words from $e$, the pre-trained sentiment model $M$ gives another  label ${L}_{e-D}$ to the remaining words $e-D$. 

\begin{table*}[!ht]
\scriptsize
\begin{subtable}{1\textwidth}
\centering
\begin{tabular}{|p{16cm}|}\hline
(1) {\color{Purple}clean updated room}. {\color{blue} friendly efficient staff }. {\color{YellowOrange} rate was too high 199 plus they charged 10 day for internet access in the room }.\\\hline
(2) {\color{red}the location is fantastic}.  the {\color{blue}staff are helpful and service oriented} . {\color{Purple}sleeping rooms meeting rooms and public lavatories not cleaned on a daily basis }. {\color{Purple} the hotel seems a bit old and a bit tired overall} . trolley noise outside can go into the wee hours . if you get a {\color{YellowOrange}great price} for a few nights this hotel may be a {\color{YellowOrange} good choice} . breakfast is very nice remember if you just stick to the cold buffet {\color{YellowOrange} it is cheaper} .\\\hline
(3) {\color{red} location is nice} . but goes from bad to worse once you walk through the door . {\color{blue} staff very surly and unhelpful }.   {\color{Purple}room and hallway had a very strange smell . rooms very run down }. {\color{Purple}  so bad that i checked out immediately and went to another hotel }. {\color{blue}  intercontinental chain should be ashamed }.\\\hline
(4) i took my daughter and her step sister to see a show at webster hall .  {\color{YellowOrange}it is so overpriced i 'm in awe} . i felt safe .  {\color{Purple}the rooms were tiny} .  {\color{red} lots of street noise all night from the partiers at the ale house below }. \\\hline
\end{tabular}
 \caption{Examples of minimal set of erased words based on {\it Bi-LSTM} model}
 \end{subtable}
~~\\~~\\
\begin{subtable}{1\textwidth}
\centering
\begin{tabular}{|p{16cm}|}\hline
(1) {\color{Purple}clean updated room}. {\color{blue} friendly efficient staff }. {\color{YellowOrange} rate was too high 199 }plus they charged 10 day for internet access in the room .\\\hline
~the {\color{red} location is fantastic}. the {\color{blue}staff are helpful and service oriented} . 
(2) {\color{Purple}sleeping rooms meeting rooms and public lavatories not cleaned} on a daily basis . the hotel seems a bit old and a bit tired overall . trolley noise outside can go into the wee hours . if you get a {\color{YellowOrange}great price} for a few nights this hotel may be a {\color{YellowOrange} good choice} . breakfast is very nice remember if you just stick to the cold buffet {\color{YellowOrange} it is cheaper} .\\\hline
(3) {\color{red} location is nice} . but goes from bad to worse once you walk through the door . {\color{blue} staff very surly and unhelpful }.   {\color{Purple}room and hallway had a very strange smell . rooms very run down }. so bad that i checked out immediately and went to another hotel . intercontinental chain should be ashamed .\\\hline
(4) i took my daughter and her step sister to see a show at webster hall .  {\color{YellowOrange}it is so overpriced i 'm in awe} . i felt safe .  {\color{Purple}the rooms were tiny} .  {\color{red} lots of street noise} all night from the partiers at the ale house below . \\\hline
\end{tabular}
 \caption{Examples of minimal set of erased words based on {\it memory-network} model.}
 \end{subtable}

\caption{Examples of minimal set of erased words to change the model decision for different aspects based on different models. Each of the colors represents a specific aspect, i.e., {\color{Purple} rooms},  {\color{blue}service},  {\color{YellowOrange}value} and    {\color{red}location}.}
\label{RL}
\end{table*}

To train the policy model, a reward function is necessary. The policy model receives a reward of $1$ if the label is changed, i.e., ${L}_{e-D}^*\neq {L}_{e}$, and $0$  if the label remains the same. 
Since we not only want the label to be changed, but also want to find the minimal set of words to change the label, the reward is scaled by the number of the words that are removed. This means removing more words will be rewarded less than removing fewer words if both of them the change the classification label.  We therefore propose the following reward:
\begin{equation}
L(e, D)=\frac{1}{|D|}\cdot {\bf 1}( {L}_{e-D}\neq {L}_{e})
\end{equation}
We also add a regularizer that encourages similar values of $z$ for words within the same sentence to encourage (or discourage) leaving out contiguous phrases:
\begin{equation}
\Omega(e,z)=\gamma\sum_{s\in S}\sum_{t\in s}|z_t-z_{t-1}|
\end{equation}
where $S$ denotes the collection of sentences by breaking the input $e$. 
Such an idea is inspired by group lasso  \cite{meier2008group}, which has been widely employed in many NLP tasks, such as document classification \cite{yogatama2014linguistic} and providing rationales for neural model interpretation
\cite{lei2016rationalizing}.  
The final reward is then:

\begin{equation}
R(e)=L(e,D)-\Omega(z_{1:N})
\end{equation}
The system is trained to maximize the expected reward of the sequence of erasing/not-erasing decisions:
  \begin{equation}
 J(\theta)=\mathbb{E}_{\pi}(R(e)|\theta)
 \label{lb1}
 \end{equation}
 The gradient of \eqref{lb1} is approximated using the likelihood ratio trick \cite{williams1992simple,glynn1990likelihood,aleksand1968stochastic}, in which for a given $e$, 
  we sample a sequence of decisions based on $\pi$, compute the associated reward and backward propagate gradients to update $\pi$, which can be summarized as follows:
 \begin{multline}
\nabla J(\theta)\approx \nabla\log\cdot\pi(z_{1:N}|\theta) (R(e)-b(e)) 
\label{action}
\end{multline}
Here $b(e)$ denotes a baseline value, to reduce the variance of the estimate while keeping it unbiased.\footnote{
 To estimate the baseline value, 
 we train another neural network model  to estimate the reward of input $e$ under current policy $\pi$, similar to \newcite{ranzato2015sequence}. } 

The policy model is trained to interpret the pre-trained sentiment classification model. Therefore, during the RL training, the original sentiment model is kept fixed. 
\paragraph{Task, Dataset and Training} 
Inspired by recent visualization work from \newcite{lei2016rationalizing},
we focus on 
the task of document-level aspect rating prediction   \cite{tang2015document,tang2015user}.
We collected  hotel reviews 
 from TripAdvisor.  
  The dataset  contains roughly 870,000 reviews with an average length of 120 words. 
 Each review contains ranking scores (integers from 1 to 5) for different aspects of the hotel, such as service, cleanliness, 
location, rooms, etc. 
We choose the aspect sentiment classification task because each review might contain diverse sentiments towards different aspects, and it is interesting to see how a model manages or fails to identify these different aspects and their associated scores when entangled with other aspects. We focus on four aspects: value,  
rooms,
service and location. 

Since
the sentiment correlation between any pair
of aspects (and the overall score) is high, the result of which may confuse the model, we employ a strategy similar to that of \newcite{lei2016rationalizing} to pick less correlated examples. 
For a given aspect, we pick the 50,000 reviews for which the score of this aspect deviates the most from the mean of the other aspects. 
We use two different models
to map input reviews to vector representations:
a vanilla Bi-LSTM and a memory-network structure \cite{sukhbaatar2015end} similar to \newcite{tang2016aspect} with attention at both word level and sentence level.
Model accuracies are shown in Appendix Table \ref{AspectResult}.

The representation is then fed to a 5-class softmax function. 
Given a trained $M$, we then train (with RL) a policy to discover the minimal set of words to erase to flip the model's classification decision. 

\subsection{Results}
Sample results are presented in Table~\ref{RL}. The reinforcement learning model 
identifies aspect-specific sentiment phrases, providing a rationale for why the sentiment model makes a certain decision. 
 By comparing Table~\ref{RL}a with Table~\ref{RL}b, we can see that the
reinforcement model trained based on the
 memory-based model offers better interpretability than the one trained based on {\it LSTMs}.
The latter model 
not only requires erasing more words to flip the model's decision,
but also sometimes deletes passages describing different aspects or overall sentiment. 
Since the RL model is trained based on the representations outputted from the sentiment model, 
better interpretability of the RL model indicates the superiority of the  memory-based
sentiment model.    
\section{Conclusion}
In this paper, we propose a general methodology for interpreting 
neural network 
decisions by analyzing the effect of erasing particular representations. 
By analyzing the harm this erasure does, 
the proposed framework offers many interpretable explanations for various aspects of neural models; 
by analyzing the benefit this erasure introduces, namely, the cases in which the removal of a representation actually improves a model's decision, the framework provides a way to conduct error analysis on neural model decisions, which has the potential the benefit a wide variety of models and tasks.

\bibliographystyle{acl2012}
\bibliography{messup}

\section{Appendix}
\subsection{Dataset Statistics and Training Accuracy for Feature Classification in Section 3}
\begin{table}[!ht]
\small
\centering
\begin{tabular}{l@{\hspace{2ex}}c@{\hspace{2ex}}c@{\hspace{2ex}}c@{\hspace{2ex}}c}
Task&$\#$Training&$\#$Dev&$\#$Test&$\#$Class \\\hline
POS&875,462&126,419&124,202&45\\
NER&189,403&47,959&42,723&6\\
Chunk&203,359&20,336&45,470&3\\
Prefix&41,406&4,601&5,076&250\\
Suffix&63,946&7,106&7,752&250\\
Sentiment&4,950&551&446&3\\
 Shape&89,864&8,987&10,126&22\\
 Frequency&123,235&13,693&15,050&--\\\hline
\end{tabular}
\caption{Statistics of datasets for dimension visualization tasks.}
\label{Statistics-task}
\end{table}
\begin{table}[!ht]
\centering
\small
\begin{tabular}{ccccc}
Aspect&service&location&rooms&value\\\hline
SVM+Uni&40.1&53.8&42.0&39.0 \\
SVM+Bi&43.2&53.1&41.1&46.1 \\
Bi-LSTM&37.5&51.4&29.8&30.5 \\
Tang (2016)&43.2&54.0&39.4&38.0\\\hline
\end{tabular}
\caption{Results for aspect rating classification (5-class) from different models. }
\label{AspectResult}
\end{table}

\begin{itemize}
\item POS Tagging: 
Each word is associated with a unique tag that indicates its syntactic role, such as plural noun, adverb, etc.
We follow the standard Penn Treebank split, using sections 0-18/19-21/22-24 as training/dev/test sets, respectively.
\item NER Tagging: 
Each word is associated with a named entity tag, such as ``person'' or ``location''. 
We evaluate on the CoNLL-2003 shared benchmark dataset for NER
 \cite{tjong2003introduction}.
 \item Chunking: 
  Each word is assigned only one unique
tag, encoded as a begin-chunk (e.g. B-NP) or inside-chunk tag (e.g. I-NP).
 We use the CoNLL-2000 dataset, in which
sections 15-18 of WSJ data are
used for training and section 20 for testing. Validation is performed by splitting the training
set.
\end{itemize}
\begin{itemize}
\item Prefix and Suffix: Words are segmented using the Morfessor package \cite{creutz2007unsupervised}. We retained top 250 frequent prefixes and suffixes. Other than ``s'' and numbers, single characters are abandoned.
We kept a list of 200,000 most frequent words, 51,083 of which are matched with a prefix and 78,804 of which are matched with a suffix. We split words into train/dev/test splits in the ratio 0.8/0.1/0.1. 
\item Sentiment: We use the MPQA subjectivity lexicon list \cite{deng2015mpqa,wilson2005recognizing}, which consists roughly 8,000 lexicons. 
\item Word shape: words are mapped to X, XX, XXX, etc. based on the number characters it contains.
\item Word frequency: the number of word occurrences is computed using a Wikipedia dump and is then mapped to log space.  Unlike all the others, which are multi-class classification tasks, word-frequency prediction is a regression task: minimize the mean squared error predicting the log frequency of each word. 
\end{itemize}

\begin{table*}[!ht]
\small
\centering
\begin{tabular}{l@{\hspace{2ex}}l@{\hspace{2ex}}c@{\hspace{2ex}}c@{\hspace{2ex}}c@{\hspace{2ex}}c@{\hspace{2ex}}c@{\hspace{2ex}}c@{\hspace{2ex}}c@{\hspace{2ex}}c@{\hspace{2ex}}c}
Training Strategy& Vector&POS&NER&Chunk&Prefix&Suffix&Sentiment&Shape&Freq \\\hline
Vanilla (Figure \ref{dimension-result}c)&GloVe&0.912&{ 0.954}&0.921&0.334&0.208&0.857&0.256&{\bf 0.349} \\
d31 removed (Figure \ref{dimension-result}d)& GloVe&{\bf 0.915}&{ 0.954}&0.921&0.336&0.207&0.818&{\bf 0.259} &0.355\\
d31, d26 removed (Figure \ref{dimension-result}e)& GloVe&0.914&{\bf 0.959}& {\bf 0.923}&{\bf 0.339}&{ 0.209}&{\bf 0.860}&0.250&0.413\\
Dropout 0.2  (Figure \ref{dimension-result}f)&GloVe&0.857&0.953&0.907&0.317&{\bf 0.239}&0.820&0.240&0.861\\\hline
Vanilla (Figure \ref{dimension-result}a)&word2vec &0.911&{ 0.954}&0.918&0.301&0.161&0.826&0.236&1.059 \\
Dropout 0.2 (Figure \ref{dimension-result}a)& word2vec & 0.889 &0.952 &0.893&0.289&0.154&0.819&0.224&1.486 \\\hline
\end{tabular}
\caption{Testing accuracy for different training strategies on tagging tasks.}
\label{test-acc-dimension}
\end{table*}

A summary of the datasets is given in Table \ref{Statistics-task}. 
Test accuracy/error for different training strategies presented in Figure \ref{dimension-result} are shown in Table \ref{test-acc-dimension}.
For classification tasks (i.e., POS, NER, Chunking, Prefix, Suffix, Sentiment, Word Shape), we report
accuracy; higher values of accuracy are better. 
For the regression task (Frequency), we report the Mean Squared Loss (loss for short); lower values of loss are better.

\subsection{Stanford Sentiment Treebank and Training Detail}
The
Stanford Sentiment Treebank is a benchmark
dataset widely used for neural model evaluations.
The dataset contains gold-standard sentiment labels
for every parse tree constituent, from sentences to
phrases to individual words, for a total of 215,154 phrases
in 11,855 sentences. The task is to perform both
fine-grained (very positive, positive, neutral, negative
and very negative) and coarse-grained (positive
vs. negative) classification at both the phrase and
sentence level.

\subsection{Aspect Rating Prediction}
The results for aspect rating prediction using the two models along with other baselines are shown in Table \ref{AspectResult}.
Feature based SVM models are trained using SVM-light package \cite{joachims2002learning}. 
LSTM based models do not perform as competitively as simple bigram-based classification models in aspect classification tasks, which has also been observed in \newcite{tang2016aspect}.

\begin{figure*}[!ht]
\centering
\includegraphics[width=3in]{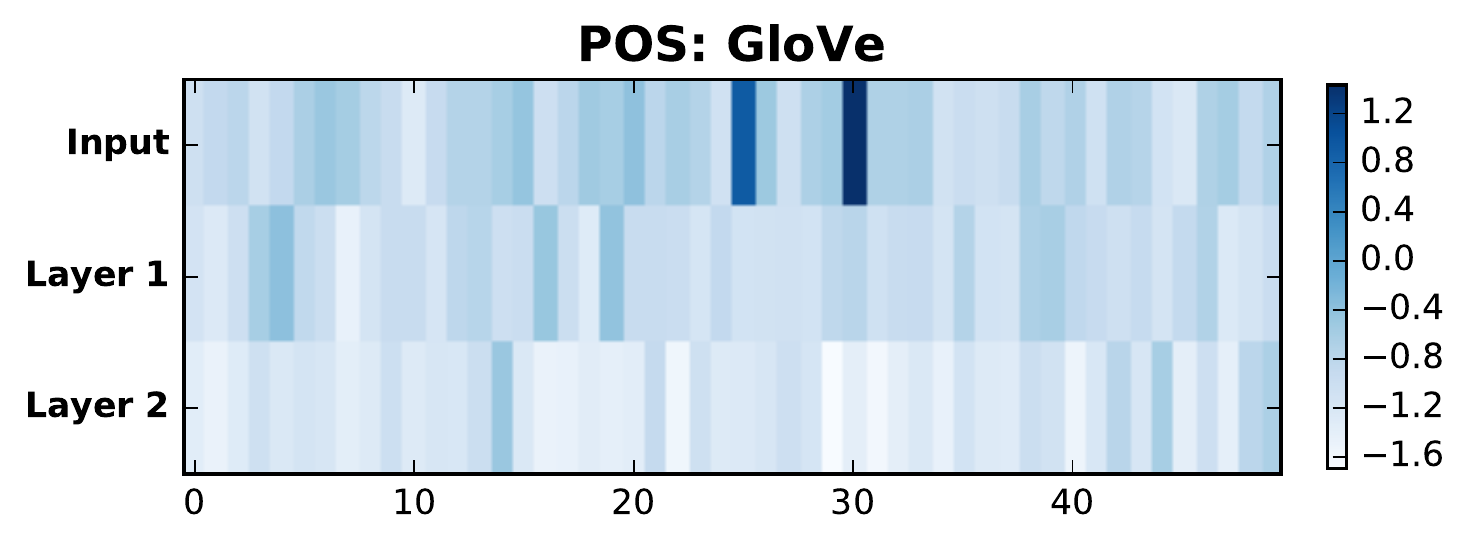}
\includegraphics[width=3in]{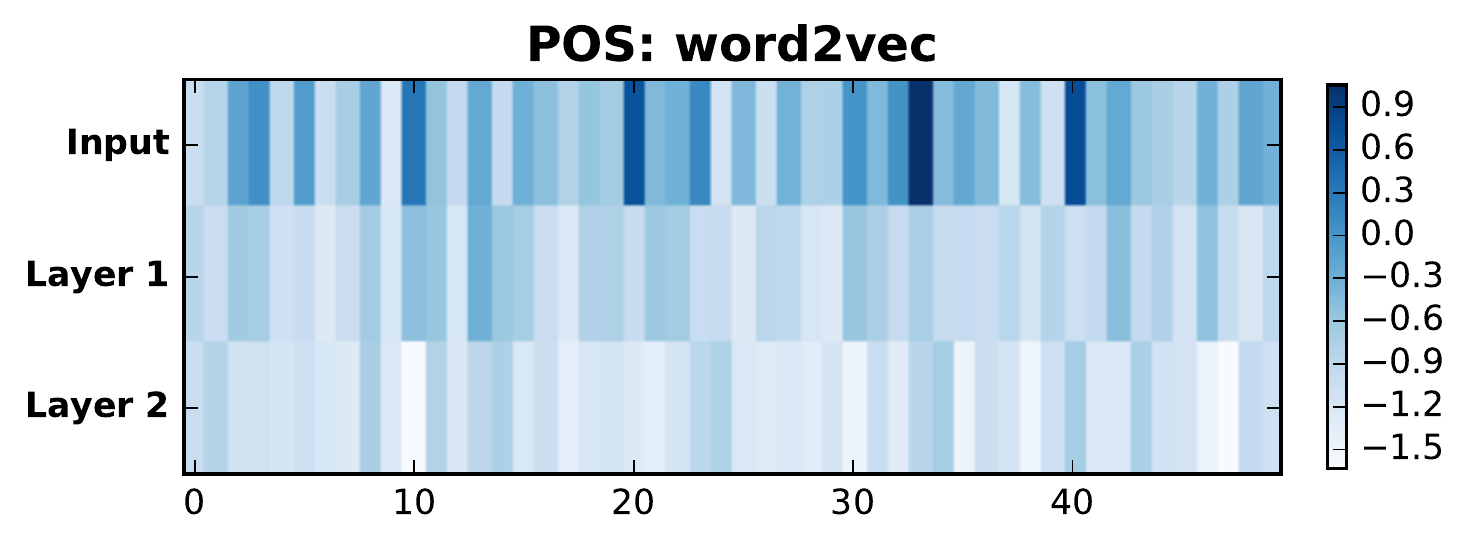}
\includegraphics[width=3in]{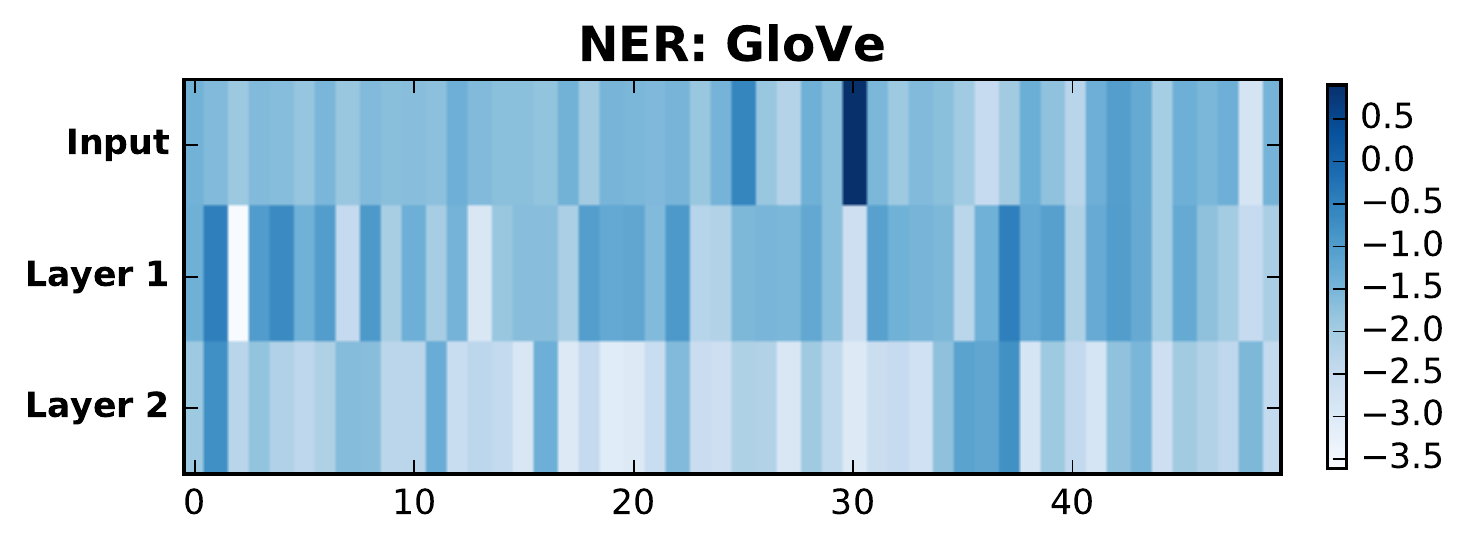}
\includegraphics[width=3in]{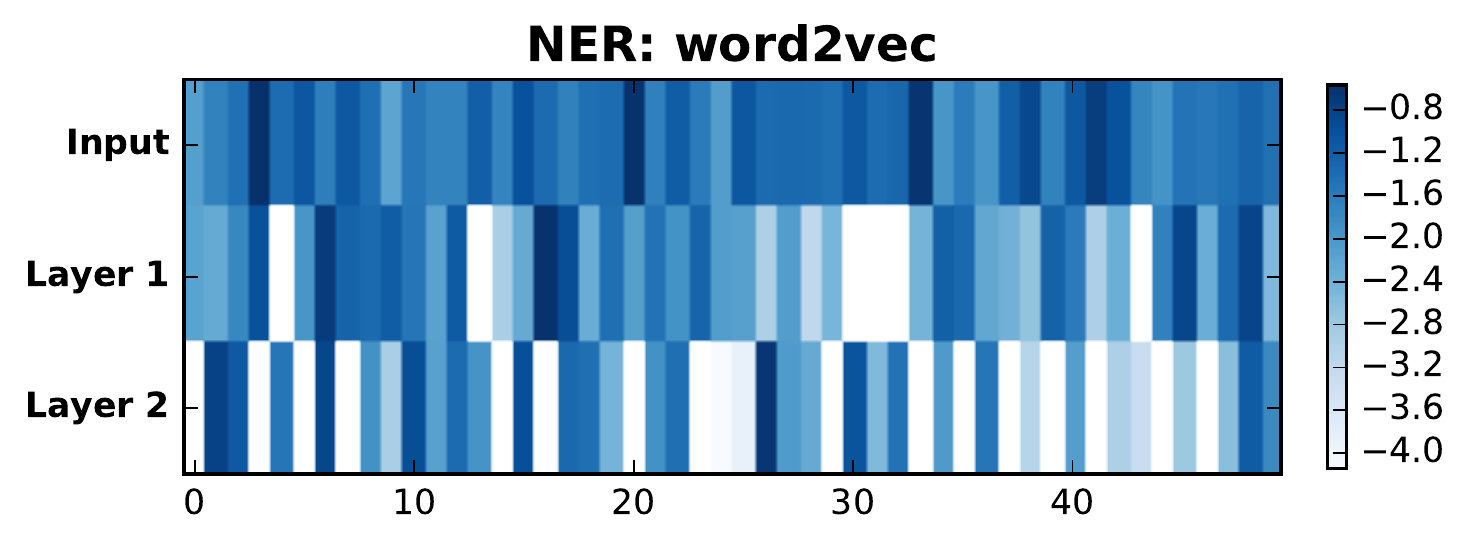}
\includegraphics[width=3in]{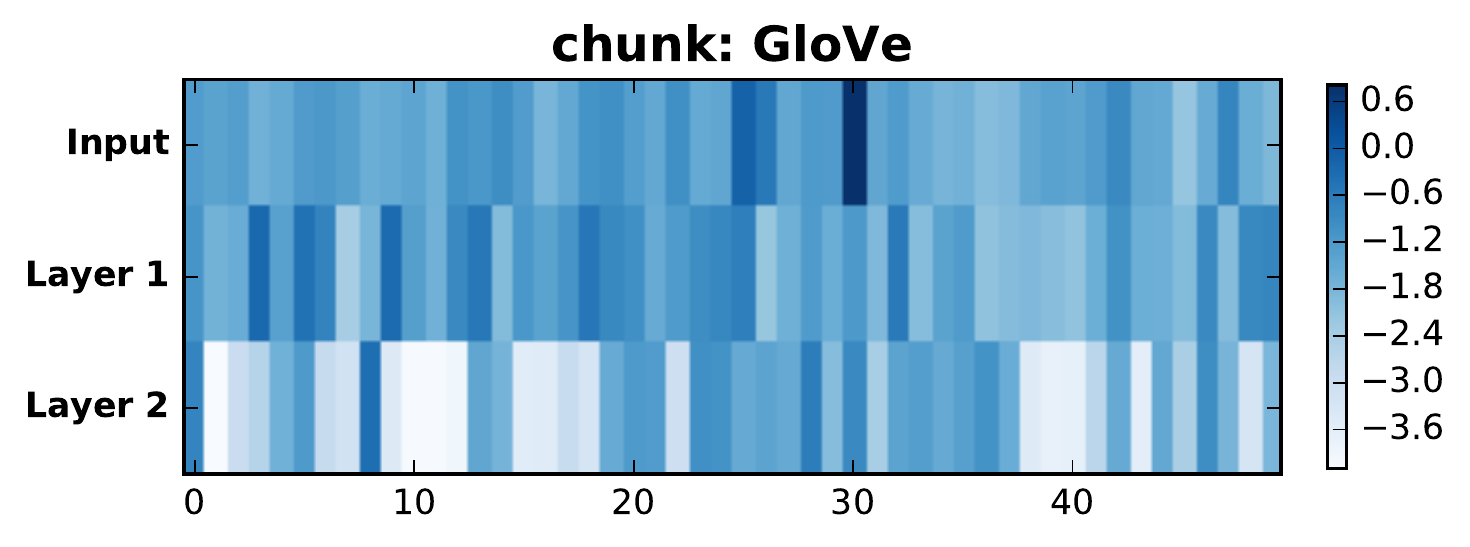}
\includegraphics[width=3in]{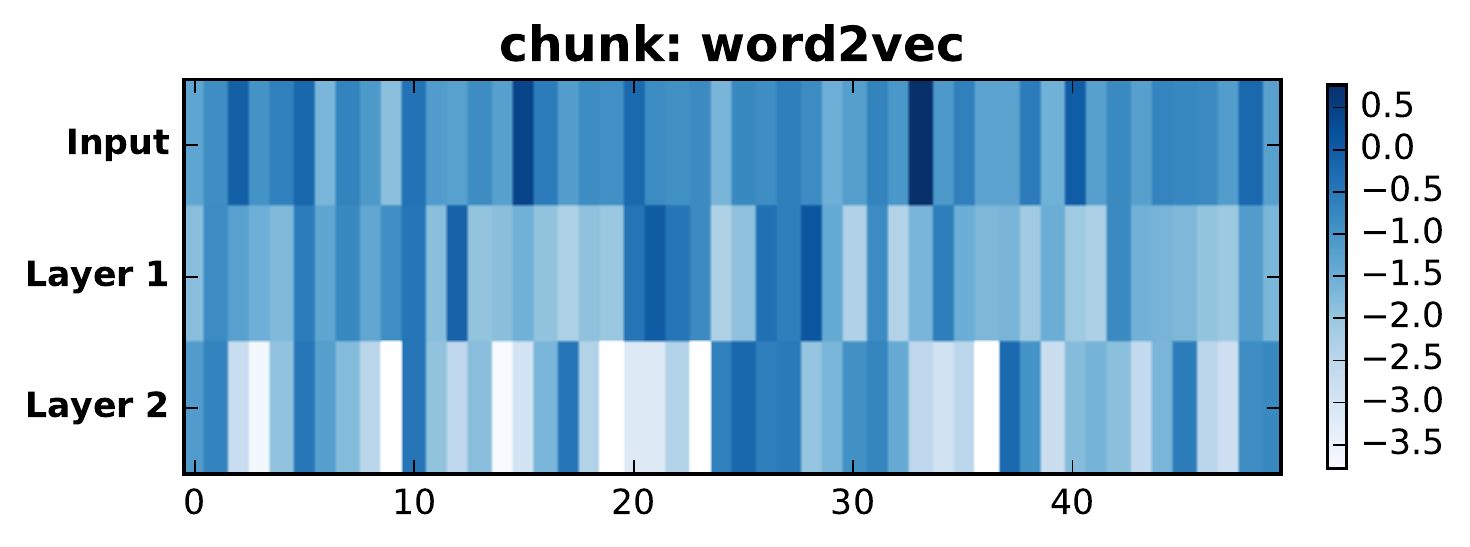}
\includegraphics[width=3in]{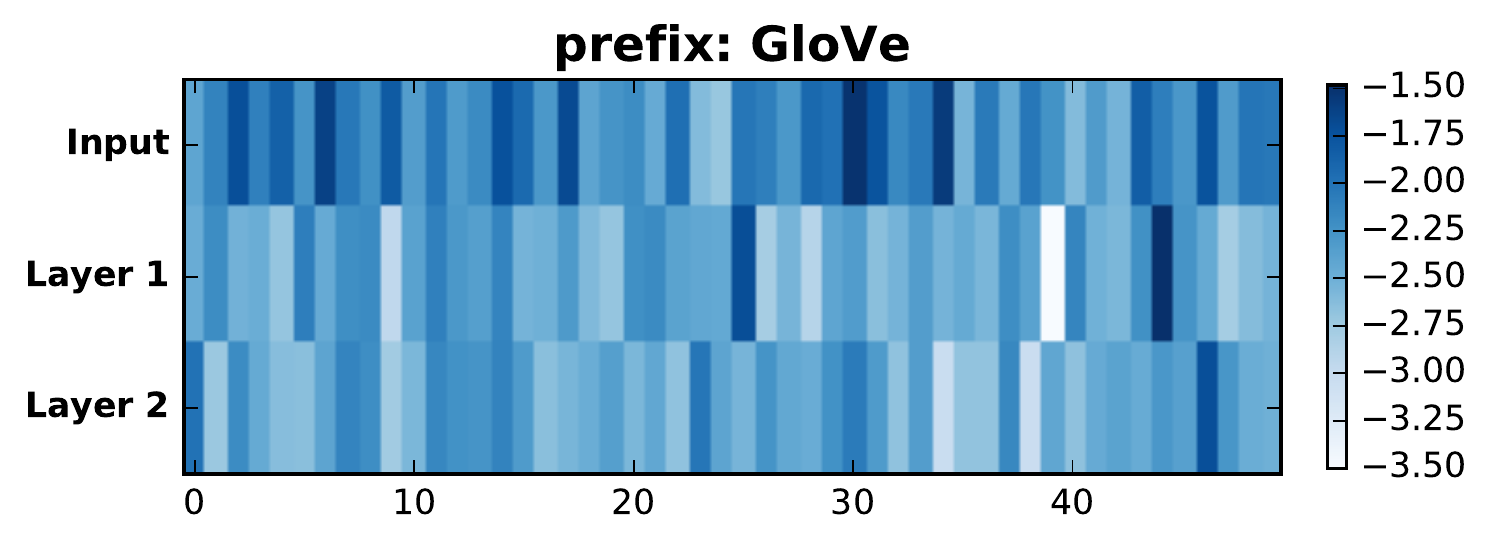}
\includegraphics[width=3in]{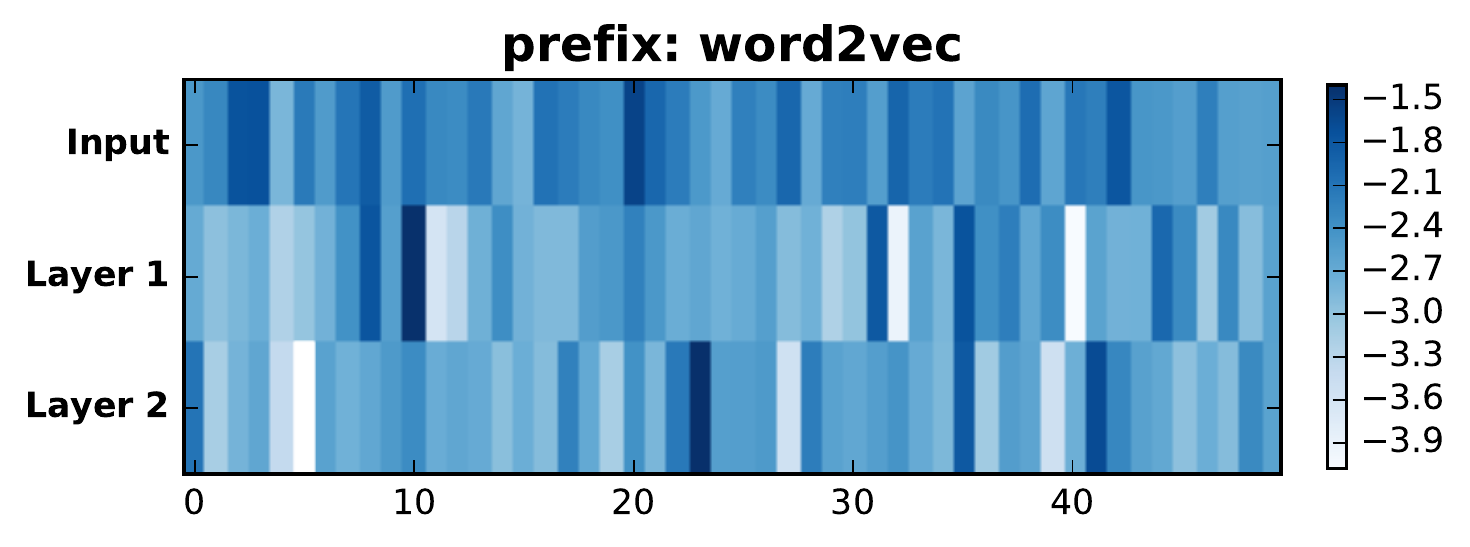}
\includegraphics[width=3in]{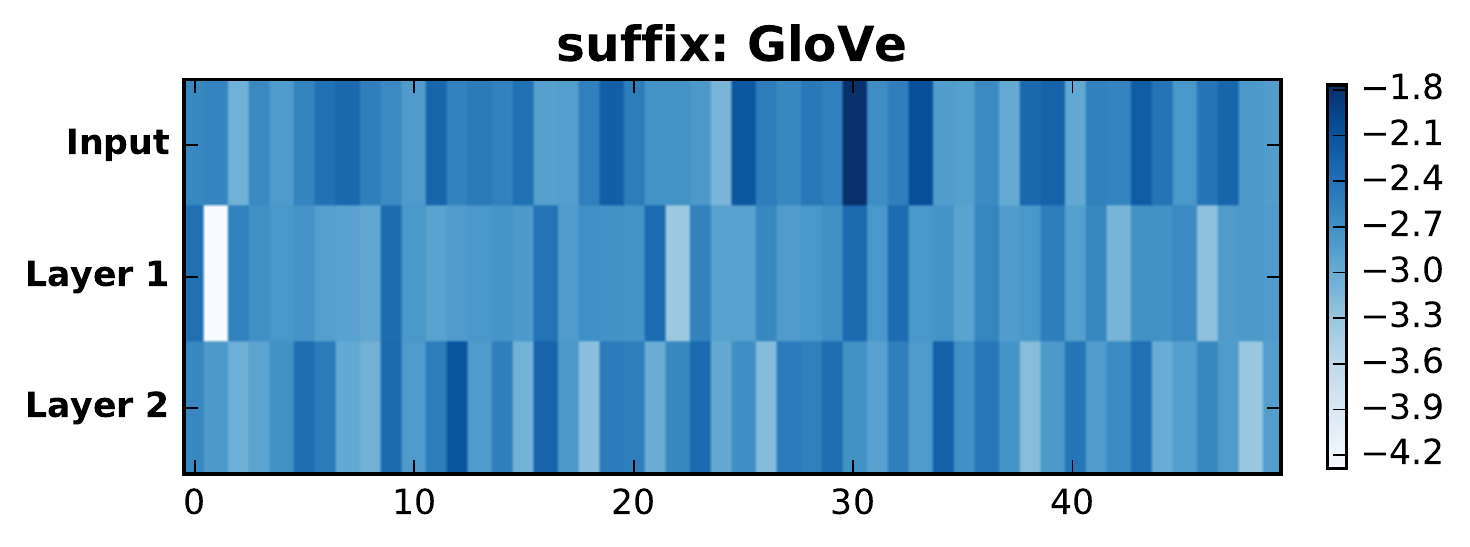}
\includegraphics[width=3in]{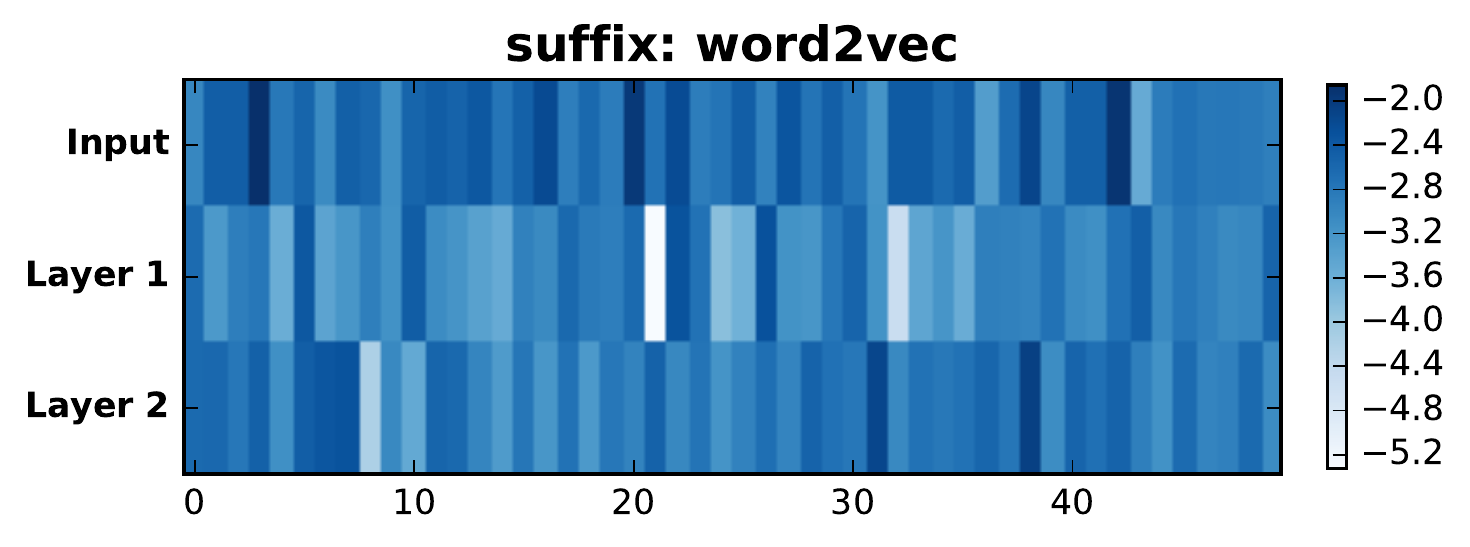}
\includegraphics[width=3in]{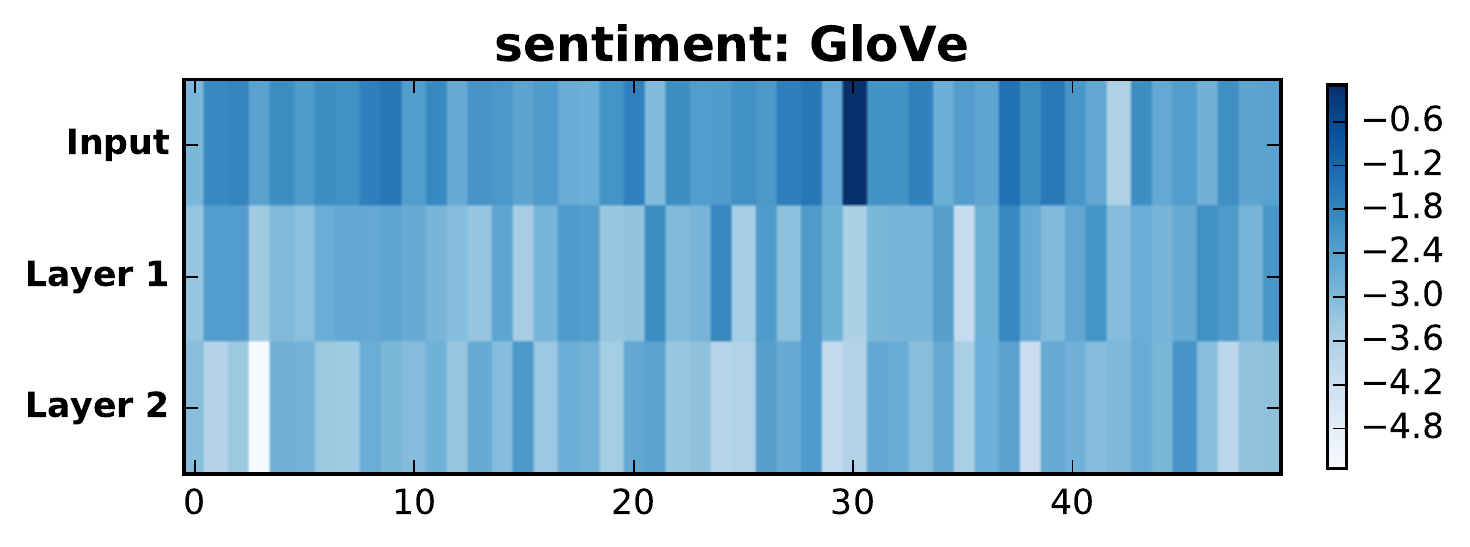}
\includegraphics[width=3in]{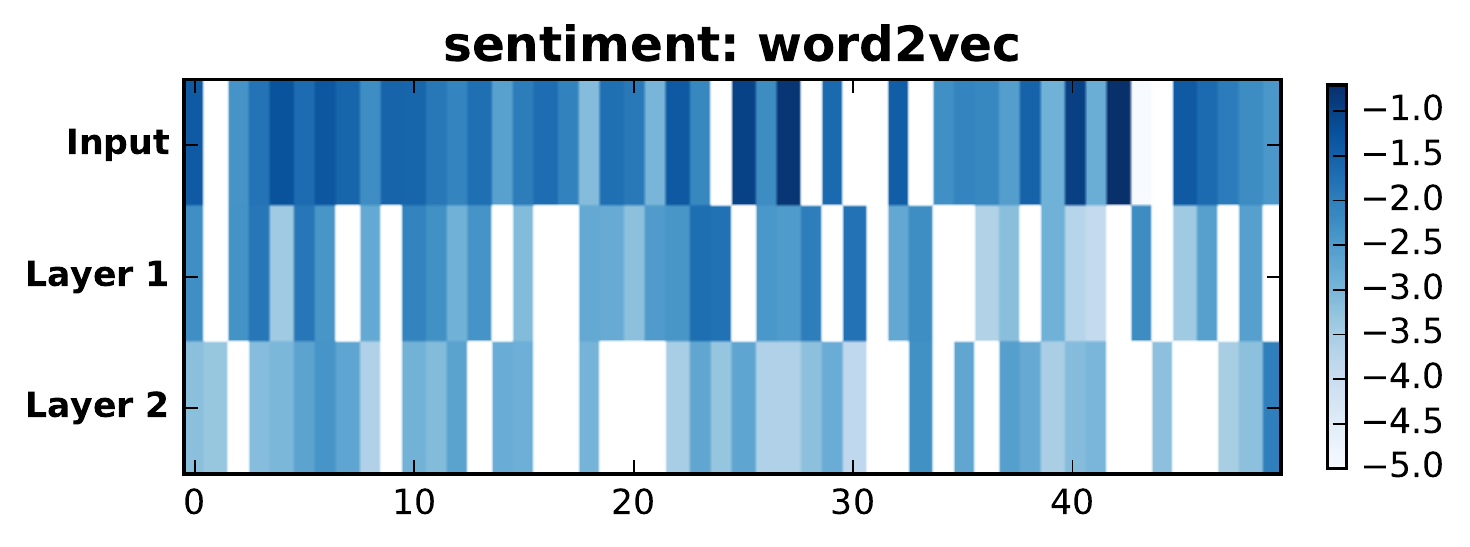}
\includegraphics[width=3in]{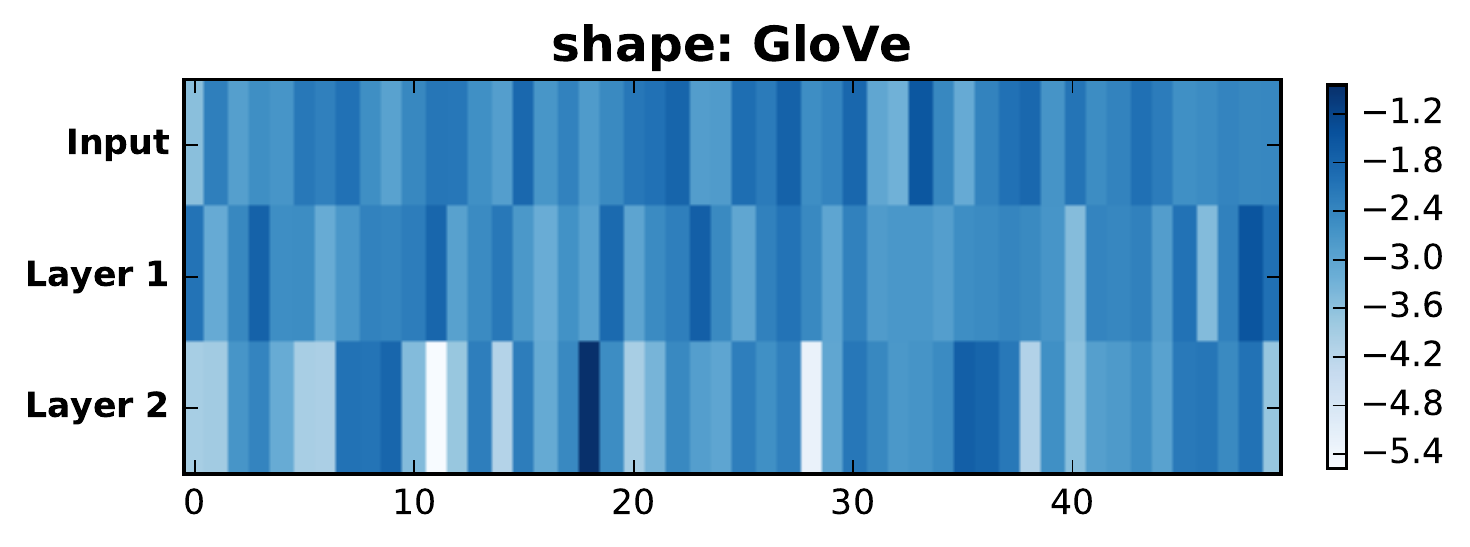}
\includegraphics[width=3in]{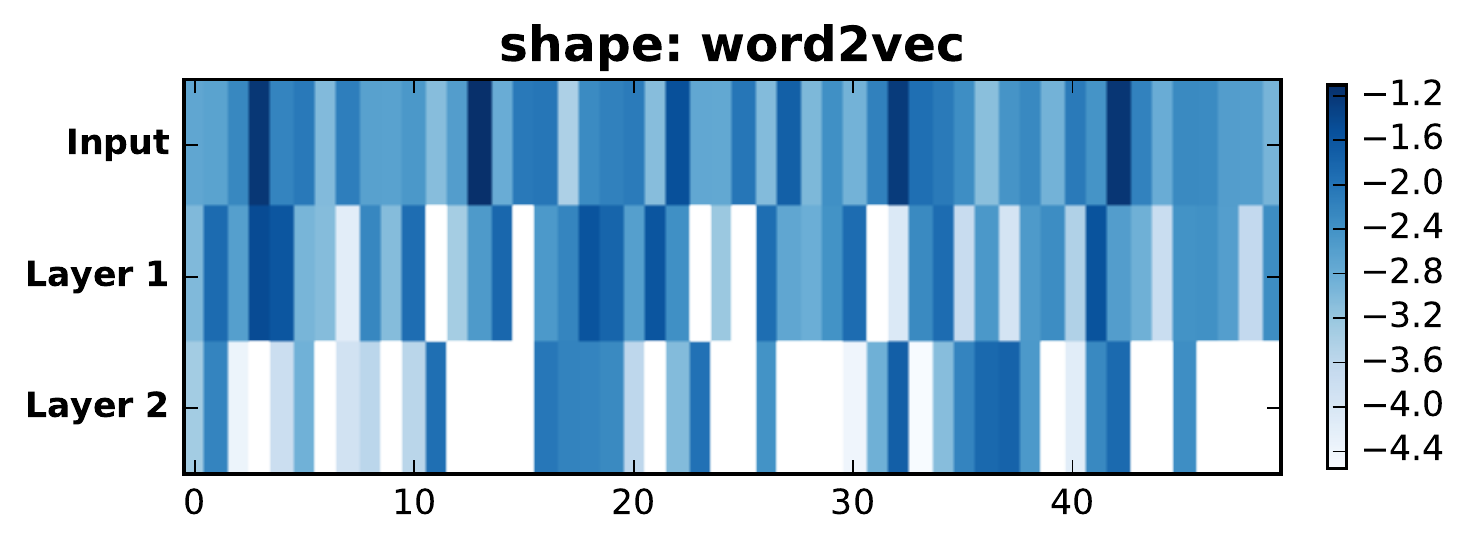}
\includegraphics[width=3in]{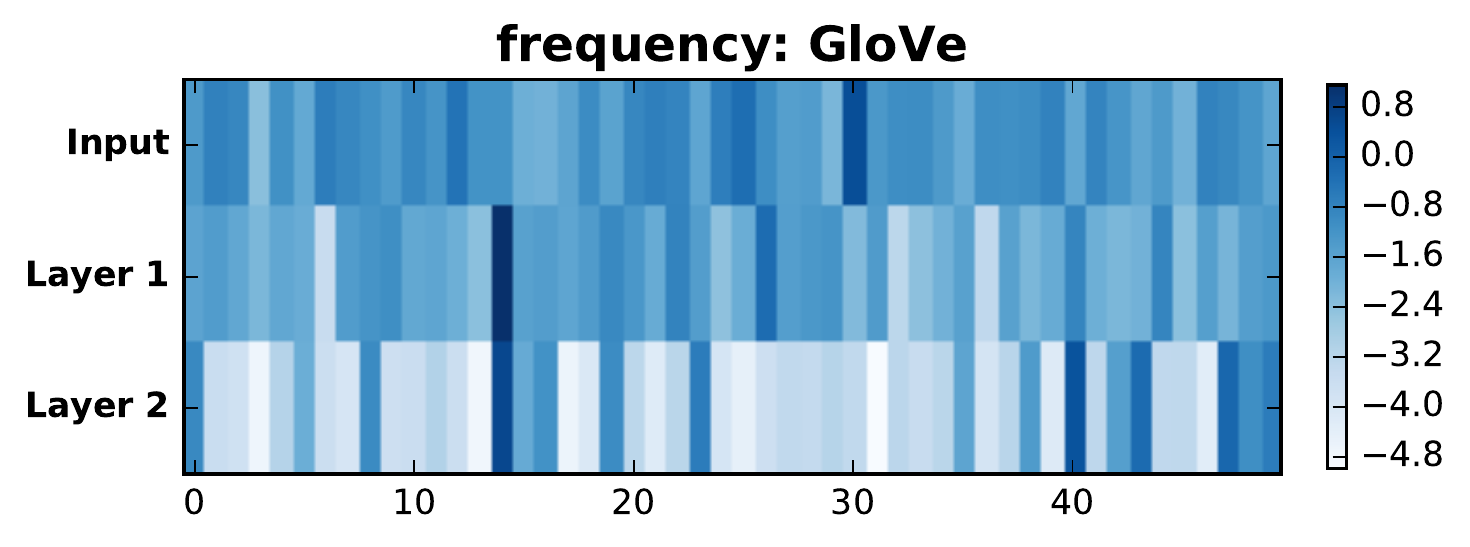}
\includegraphics[width=3in]{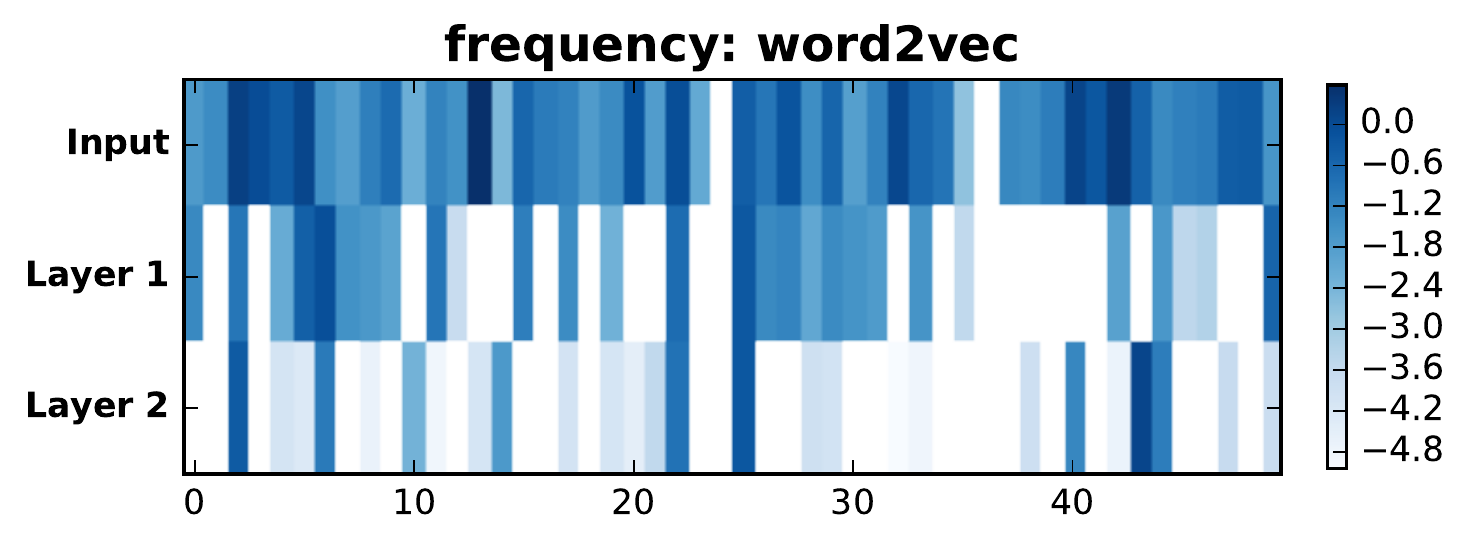}
\caption{Heatmap of importance  (computed using Eq.~\ref{di}) of each layer for different tasks. 
Each column denotes a dimension and each row denotes a layer in the network. 
Importance values are transformed to log space.
}
\label{layer_all_tasks}
\end{figure*}

\begin{table*}[!ht]
\centering
\small		
\begin{tabular}{cccc}
rank&bi-lstm&uni-lstm&rnn\\\hline
11&wonderful (9.529)&stillborn (5.939)&flawless (3.250) \\ 
12&bastard (9.464)&pleasurably (5.857)&heart-stopping (3.231) \\ 
13&greatest (9.460)&savor (5.854)&unwatchable (3.204) \\ 
14&brilliant (8.350)&succeeds (5.767)&tremendous (2.832) \\ 
15&worst (7.739)&punch-drunk (5.728)&lukewarm (2.820) \\ 
16&excellent (6.835)&inert (5.594)&cop-out (2.684) \\ 
17&dumbness (6.633)&greatest (5.593)&drab (2.654) \\ 
18&nicely (6.605)&irresistible (5.461)&incredible (2.419) \\ 
19&hated (6.557)&nicely (5.444)&sweetest (2.365) \\ 
20&lukewarm (6.443)&brilliant (5.361)&waste (2.342) \\ 
21&unpleasant (6.355)&skillfully (5.158)&overstuffed (2.320) \\ 
22&clunker (5.754)&must-see (5.125)&vulgar (2.312) \\ 
23&cop-out (5.712)&excellent (4.883)&lackluster (2.269) \\ 
24&beaut (5.650)&bothersome (4.728)&bothersome (2.256) \\ 
25&beautiful (5.401)&worst (4.698)&punish (2.119) \\ 
26&must-see (5.362)&heart-stopping (4.347)&pleasurably (2.085) \\ 
27&deliciously (5.226)&refreshing (4.225)&muted (1.863) \\ 
28&sabotages (5.103)&invigorating (4.137)&excellent (1.859) \\ 
29&irresistible (4.977)&travesty (3.977)&fabulous (1.853) \\ 
30&best (4.916)&fabulous (3.949)&dumbness (1.834) \\ 
31&incredible (4.914)&eye-popping (3.919)&stupidest (1.777) \\ 
32&stupider (4.810)&incredible (3.875)&flaccid (1.768) \\ 
33&fabulous (4.585)&imaginative (3.823)&clunker (1.756) \\ 
34&waste (4.579)&deliciously (3.802)&ridiculous (1.717) \\ 
35&disappointingly (4.565)&misses (3.722)&suffocated (1.653) \\ 
36&tremendous (4.302)&incarnates (3.694)&sorry (1.624) \\ 
37&bothersome (4.212)&waste (3.600)&jackasses (1.579) \\ 
38&pet (4.141)&hated (3.512)&turgid (1.544) \\ 
39&misses (4.139)&feast (3.442)&snoozer (1.515) \\ 
40&wannabe (4.072)&snoozer (3.368)&unforgettable (1.482) \\ 
41&repulsive (4.071)&disaster (3.362)&deliciously (1.418) \\ 
42&bracing (4.043)&wonderful (3.298)&blandness (1.399) \\ 
43&ingenious (4.019)&stupider (3.243)&delightful (1.337) \\ 
44&moot (3.981)&clunker (3.231)&exasperating (1.320) \\ 
45&invigorating (3.940)&mesmerizing (3.185)&failed (1.317) \\ 
46&snoozer (3.928)&lukewarm (3.163)&roller-coaster (1.287) \\ 
47&punch-drunk (3.877)&rent (3.120)&suffer (1.264) \\ 
48&overstuffed (3.831)&worthwhile (3.034)&achievement (1.224) \\ 
49&unwatchable (3.785)&superior (2.964)&nowhere (1.175) \\ 
50&delight (3.766)&letdown (2.959)&remarkable (1.168) \\ 
51&breezes (3.747)&hollow (2.925)&breathtaking (1.162) \\ 
52&joyful (3.734)&screenwriters (2.862)&beaut (1.161) \\ 
53&disaster (3.728)&ugliest (2.816)&awfully (1.135) \\ 
54&ungainly (3.710)&moot (2.804)&sour (1.125) \\ 
55&pleasurably (3.710)&astute (2.767)&hilarious (1.124) \\ 
56&exquisitely (3.633)&thoughtful (2.757)&monotonous (1.110) \\ 
57&marvellous (3.587)&vile (2.734)&inviting (1.104) \\ 
58&hilarious (3.533)&repulsive (2.721)&treat (1.085) \\ 
59&travesty (3.500)&likeable (2.693)&worthwhile (1.066) \\ 
60&sparkles (3.469)&bravely (2.691)&mesmerizing (1.055) \\ \hline
\end{tabular}
\caption{Top ranked words by importance  (computed using Eq.~\ref{di}) from the Uni-LSTM, Bi-LSTM and standard RNN models.}
\label{all-word-importance}
\end{table*}

\begin{table*}[!ht]
\scriptsize
\begin{tabular}{p{0.7cm}p{2cm}p{1cm}p{1cm}p{10cm}}\\
rank&Word&Score&Label&Original Sentence \\\hline
1&revelatory&-0.90&-&flat, but with a revelatory performance by michelle williams.\\
2&lacks&-0.88&+&what it lacks in originality it makes up for in intelligence and b-grade stylishness.\\
3&shame&-0.84&+&it takes this never-ending confusion and hatred, puts a human face on it, evokes shame among all who are party to it and even promotes understanding.\\
4&skip&-0.83&+&skip work to see it at the first opportunity.\\
5&lackadaisical&-0.82&+&a pleasant ramble through the sort of idoosyncratic terrain that errol morris has often dealt with... it does possess a loose, lackadaisical charm.\\
6&by-the-books&-0.82&+&a fairly by-the-books blend of action and romance with sprinklings of intentional and unintentional comedy.\\
7&misses&-0.82&++&this is cool, slick stuff, ready to quench the thirst of an audience that misses the summer blockbusters.\\
8&bonehead&-0.82&++&the smartest bonehead comedy of the summer.\\
9&dingy&-0.81&+&it's a nicely detailed world of pawns, bishops and kings, of wagers in dingy backrooms or pristine forests.\\
10&enjoying&-0.81&-&i kept thinking over and over again,' i should be enjoying this.'\\
11&foul&-0.80&+&a whole lot foul, freaky and funny.\\
12&best&-0.80&-&the best way to hope for any chance of enjoying this film is by lowering your expectations.\\
13&confident&-0.79&-&just when the movie seems confident enough to handle subtlety, it dives into soapy bathos.\\
14&inconsequential&-0.77&+&has a shambling charm... a cheerfully inconsequential diversion.\\
15&oblivion&-0.77&++&... mesmerizing, an eye-opening tour of modern beijing culture in a journey of rebellion, retreat into oblivion and return.\\
16&captivating&-0.76&+&a captivating cross-cultural comedy of manners.\\
17&acidic&-0.75&++&hilarious, acidic brit comedy.\\
18&overblown&-0.75&++&occasionally funny, always very colorful and enjoyably overblown in the traditional almodóvar style.\\
19&n't&-0.75&++&a sensitive and expertly acted crowd-pleaser that is n't above a little broad comedy and a few unabashedly sentimental tears.\\
20&n't&-0.75&++&a great comedy filmmaker knows great comedy need n't always make us laugh.\\
21&inconsequential&-0.75&+&has a shambling charm... a cheerfully inconsequential diversion.\\
22&entertaining&-0.74&+&sturdy, entertaining period drama... both caine and fraser have their moments.\\
23&south-of-the-border&-0.74&-&like a south-of-the-border melrose place.\\
24&not&-0.74&++&it's a good film -- not a classic, but odd, entertaining and authentic.\\
25&pleasing&-0.72&-&an intermittently pleasing but mostly routine effort.\\
26&difficult&-0.72&++&a worthy entry into a very difficult genre.\\
27&lost&-0.72&+&gets under the skin of a man who has just lost his wife.\\
28&dahmer&-0.72&++&renner's performance as dahmer is unforgettable, deeply absorbing.\\
29&by-the-numbers&-0.72&+&`` antwone fisher'' is an earnest, by-the-numbers effort by washington.\\
30&great&-0.72&-&it's a great deal of sizzle and very little steak.\\
31&exuberantly&-0.72&+&zany, exuberantly irreverent animated space adventure.\\
32&dumb&-0.71&+&the transporter is as lively and as fun as it is unapologetically dumb\\
33&fascinating&-0.71&+&it's both a necessary political work and a fascinating documentary...\\
34&insultingly&-0.71&+&it is so refreshing to see robin williams turn 180 degrees from the string of insultingly innocuous and sappy fiascoes he's been making for the last several years.\\
35&none&-0.71&+&as a witness to several greek-american weddings -- but, happily, a victim of none -- i can testify to the comparative accuracy of ms. vardalos' memories and insights.\\
36&squalor&-0.71&+&the result is mesmerizing -- filled with menace and squalor.\\
37&insurance&-0.70&-&technically, the film is about as interesting as an insurance commercial.\\
38&mess&-0.70&-&just a bloody mess.\\
39&inviting&-0.70&-&we are left with a superficial snapshot that, however engaging, is insufficiently enlightening and inviting.\\
40&comedy&-0.70&+&a pleasant romantic comedy.\\
41&bravura&-0.70&-&a bravura exercise in emptiness.\\
42&n't&-0.70&0&in the end, white oleander is n't an adaptation of a novel.\\
43&well&-0.70&-&well, it does go on forever.\\
44&unhappily&-0.70&+&happily for mr. chin -- though unhappily for his subjects -- the invisible hand of the marketplace wrote a script that no human screenwriter could have hoped to match.\\
45&slugs&-0.70&0&melanie eventually slugs the yankee.\\
46&good&-0.69&-&an ultra-low-budget indie debut that smacks more of good intentions than talent.\\
47&time-wasting&-0.69&0&... an agreeable time-wasting device -- but george pal's low-tech 1960 version still rules the epochs.\\
48&departure&-0.69&+&not for everyone, but for those with whom it will connect, it's a nice departure from standard moviegoing fare.\\
\hline
\end{tabular}
\caption{Words ranked by negative importance score  (computed using Eq.~\ref{di}) for the Bi-LSTM model. Negative importance means that the model makes better predictions when the word is erased. ++, +, 0, -, and {-}{-} respectively denote strong positive, positive, neutral, negative and strong negative  gold-standard labels from the dataset. }
\label{bi-re}
\end{table*}

\begin{table*}[!ht]
\scriptsize
\begin{tabular}{p{0.7cm}p{2cm}p{1cm}p{1cm}p{10cm}}\\
rank&Word&Score&Label&Original Sentence \\\hline
1&foul&-0.89&+&a whole lot foul, freaky and funny.\\
2&shaky&-0.84&+&as shaky as the plot is, kaufman's script is still memorable for some great one-liners.\\
3&bonehead&-0.83&++&the smartest bonehead comedy of the summer.\\
4&harsh&-0.81&+&harsh, effective documentary on life in the israeli-occupied palestinian territories.\\
5&lacks&-0.80&+&what it lacks in originality it makes up for in intelligence and b-grade stylishness.\\
6&skip&-0.79&+&skip work to see it at the first opportunity.\\
7&confident&-0.79&-&just when the movie seems confident enough to handle subtlety, it dives into soapy bathos.\\
8&shame&-0.78&+&it takes this never-ending confusion and hatred, puts a human face on it, evokes shame among all who are party to it and even promotes understanding.\\
9&claptrap&-0.78&+&more a load of enjoyable, conan-esque claptrap than the punishing, special-effects soul assaults the mummy pictures represent.\\
10&wonderful&-0.76&-&while benigni -lrb- who stars and co-wrote -rrb- seems to be having a wonderful time, he might be alone in that.\\
11&dingy&-0.74&+&it's a nicely detailed world of pawns, bishops and kings, of wagers in dingy backrooms or pristine forests.\\
12&great&-0.74&-&it's a great deal of sizzle and very little steak.\\
13&bogus&-0.73&+&-lrb- a -rrb- hollywood sheen bedevils the film from the very beginning... -lrb- but -rrb- lohman's moist, deeply emotional eyes shine through this bogus veneer...\\
14&engrossing&-0.73&-&where last time jokes flowed out of cho's life story, which provided an engrossing dramatic through line, here the comedian hides behind obviously constructed routines.\\
15&preposterous&-0.72&+&while the isle is both preposterous and thoroughly misogynistic, its vistas are incredibly beautiful to look at.\\
16&stunning&-0.71&+&hayek is stunning as frida and... a star-making project.\\
17&camouflaged&-0.71&+&a film of precious increments artfully camouflaged as everyday activities.\\
18&dumb&-0.70&+&the transporter is as lively and as fun as it is unapologetically dumb\\
19&dahmer&-0.70&++&renner's performance as dahmer is unforgettable, deeply absorbing.\\
20&disturbing&-0.70&++&disturbing and brilliant documentary.\\
21&marvelous&-0.70&+&marvelous, merry and, yes, melancholy film.\\
22&enjoyable&-0.70&+&an enjoyable film for the family, amusing and cute for both adults and kids.\\
23&thankfully&-0.69&+&farrell... thankfully manages to outshine the role and successfully plays the foil to willis's world-weary colonel.\\
24&pleasing&-0.69&-&an intermittently pleasing but mostly routine effort.\\
25&half-wit&-0.69&++&an enjoyably half-wit remake of the venerable italian comedy big deal on madonna street.\\
26&brimful&-0.69&0&brimful.\\
27&time-wasting&-0.69&0&... an agreeable time-wasting device -- but george pal's low-tech 1960 version still rules the epochs.\\
28&nothing&-0.68&+&sometimes, nothing satisfies like old-fashioned swashbuckling.\\
29&by-the-books&-0.68&+&a fairly by-the-books blend of action and romance with sprinklings of intentional and unintentional comedy.\\
30&misses&-0.68&++&this is cool, slick stuff, ready to quench the thirst of an audience that misses the summer blockbusters.\\
31&soggy&-0.67&-&a soggy, cliche-bound epic-horror yarn that ends up being even dumber than its title.\\
32&worse&-0.67&0&every so often a movie comes along that confirms one's worse fears about civilization as we know it.\\
33&no&-0.67&0&no question.\\
34&captivating&-0.67&+&a captivating cross-cultural comedy of manners.\\
35&worse&-0.66&0&it's a worse sign when you begin to envy her condition.\\
36&shambling&-0.66&+&has a shambling charm... a cheerfully inconsequential diversion.\\
37&modest&-0.65&-&a modest and messy metaphysical thriller offering more questions than answers.\\
38&captivating&-0.65&+&most of crush is a clever and captivating romantic comedy with a welcome pinch of tartness.\\
39&intelligent&-0.65&+&a mostly intelligent, engrossing and psychologically resonant suspenser.\\
40&sterile&-0.65&+&a distant, even sterile, yet compulsively watchable look at the sordid life of hogan's heroes star bob crane.\\
41&terrific&-0.65&0&the actors are so terrific at conveying their young angst, we do indeed feel for them.\\
42&unconcerned&-0.64&+&here's a british flick gleefully unconcerned with plausibility, yet just as determined to entertain you.\\
43&intriguing&-0.64&-&kwan makes the mix-and - match metaphors intriguing, while lulling us into torpor with his cultivated allergy to action.\\
44&slugs&-0.64&0&melanie eventually slugs the yankee.\\
45&south-of-the-border&-0.63&-&like a south-of-the-border melrose place.\\
46&terrific&-0.63&+&highlights are the terrific performances by christopher plummer, as the prime villain, and nathan lane as vincent crummles, the eccentric theater company manager.\\
47&mournfully&-0.63&+&noyce's film is contemplative and mournfully reflective.\\
\hline
\end{tabular}
\caption{Words ranked by negative importance score (computed using Eq.~\ref{di}) obtained by the Uni-LSTM model. Negative importance means that the model makes better prediction when the word is erased. ++, +, 0, -, and {-}{-} respectively denote strong positive, positive, neutral, negative and strong negative gold-standard sentiment labels from the dataset.}
\label{uni-re}
\end{table*}

\begin{table*}[!ht]
\scriptsize
\begin{tabular}{p{0.7cm}p{2cm}p{1cm}p{1cm}p{10cm}}\\
rank&Word&Score&Label&Original Sentence \\\hline
1&effective&-0.93&0&effective but too-tepid biopic\\
2&no&-0.89&0&no question.\\
3&high&-0.87&0&high on melodrama.\\
4&brimful&-0.86&0&brimful.\\
5&bravura&-0.85&-&a bravura exercise in emptiness.\\
6&pleasing&-0.84&-&an intermittently pleasing but mostly routine effort.\\
7&stunning&-0.83&+&hayek is stunning as frida and... a star-making project.\\
8&n't&-0.80&+&is n't it great ?\\
9&engaging&-0.80&+&an engaging overview of johnson's eccentric career.\\
10&thrill&-0.80&-&the thrill is -lrb- long -rrb- gone.\\
11&real&-0.79&{-}{-}&a real clunker.\\
12&captivating&-0.79&+&a captivating cross-cultural comedy of manners.\\
13&skip&-0.79&+&skip work to see it at the first opportunity.\\
14&insomnia&-0.78&0&insomnia is involving.\\
15&right&-0.78&0&oh, it's extreme, all right.\\
16&faultlessly&-0.78&0&faultlessly professional but finally slight.\\
17&well&-0.78&0&well before it's over, beijing bicycle begins spinning its wheels.\\
18&fun&-0.77&0&as a director, mr. ratliff wisely rejects the temptation to make fun of his subjects.\\
19&well&-0.77&-&well, it does go on forever.\\
20&absorbing&-0.77&+&an absorbing, slice-of-depression life that touches nerves and rings true.\\
21&no&-0.77&+&finally, a genre movie that delivers -- in a couple of genres, no less.\\
22&good&-0.77&-&first good, then bothersome.\\
23&harsh&-0.76&+&harsh, effective documentary on life in the israeli-occupied palestinian territories.\\
24&great&-0.76&-&it's a great deal of sizzle and very little steak.\\
25&good&-0.75&+&bullock does a good job here of working against her natural likability.\\
26&no&-0.74&+&by the end of no such thing the audience, like beatrice, has a watchful affection for the monster.\\
27&terrific&-0.73&0&the actors are so terrific at conveying their young angst, we do indeed feel for them.\\
28&best&-0.72&-&my response to the film is best described as lukewarm.\\
29&newton&-0.71&++&newton draws our attention like a magnet, and acts circles around her better known co-star, mark wahlberg.\\
30&best&-0.71&-&the best way to hope for any chance of enjoying this film is by lowering your expectations.\\
31&community&-0.71&{-}{-}&it feels like a community theater production of a great broadway play : even at its best, it will never hold a candle to the original.\\
32&hmm&-0.71&0&hmm.\\
33&invincible&-0.71&+&the invincible werner herzog is alive and well and living in la\\
34&departure&-0.71&++&greene delivers a typically solid performance in a role that is a bit of a departure from the noble characters he has played in the past, and he is matched by schweig, who carries the film on his broad, handsome shoulders.\\
35&self-aware&-0.71&-&it's fairly self-aware in its dumbness.\\
36&no&-0.70&++&waydowntown is by no means a perfect film, but its boasts a huge charm factor and smacks of originality.\\
37&happy&-0.70&0&just what makes us happy, anyway ?\\
38&vivid&-0.70&-&the essential problem in orange county is that, having created an unusually vivid set of characters worthy of its strong cast, the film flounders when it comes to giving them something to do.\\
39&understands&-0.70&0&... understands that a generation defines its music as much as the music defines a generation.\\
40&no&-0.70&+&this version's no classic like its predecessor, but its pleasures are still plentiful.\\
41&pleasant&-0.69&+&a pleasant romantic comedy.\\
42&community&-0.69&{-}{-}&it feels like a community theater production of a great broadway play : even at its best, it will never hold a candle to the original.\\
43&gimmick&-0.69&++&an endearingly offbeat romantic comedy with a great meet-cute gimmick.\\
44&community&-0.69&{-}{-}&it feels like a community theater production of a great broadway play : even at its best, it will never hold a candle to the original.\\
45&love&-0.69&0&hip-hop has a history, and it's a metaphor for this love story.\\
46&great&-0.69&{-}{-}&if melville is creatively a great whale, this film is canned tuna.\\
47&clever&-0.69&+&a clever blend of fact and fiction.\\
48&demeanor&-0.69&{-}{-}&the smug, oily demeanor that donovan adopts throughout the stupidly named pipe dream is just repulsive.\\
49&decent&-0.68&-&some decent actors inflict big damage upon their reputations.\\
50&slugs&-0.68&0&melanie eventually slugs the yankee.\\
\hline
\end{tabular}
\caption{Words ranked by negative importance score (computed using Eq.\ref{di})  obtained by the RNN model. Negative importance means that the model makes better prediction when the word is erased. ++, +, 0, -, and {-}{-} respectively denote strong positive, positive, neutral, negative and strong negative gold-standard sentiment labels from the dataset.}
\label{rnn-re}
\end{table*}

\end{document}